%% file: main.tex
\documentclass[10pt,twocolumn,letterpaper]{article}

\usepackage[pagenumbers]{cvpr} 


\definecolor{cvprblue}{rgb}{0.21,0.49,0.74}
\usepackage[pagebackref,breaklinks,colorlinks,allcolors=cvprblue]{hyperref}

\usepackage{bm}
\usepackage{longtable}
\usepackage{booktabs}   
\usepackage{tabularx}      
\usepackage{multirow} 
\usepackage{caption}
\usepackage{tabularray}
\usepackage{times}
\usepackage{epsfig}
\usepackage{makecell} 
\usepackage{graphicx}
\usepackage{amsmath}
\usepackage{amssymb}
\usepackage{multirow}
\usepackage{float}
\usepackage{booktabs}
\usepackage{pdfrender}
\usepackage{comment}
\usepackage{lipsum}  
\usepackage{xcolor}
\usepackage[table]{xcolor}
\usepackage{subcaption} 
\usepackage{siunitx}
\usepackage{comment}
\usepackage{minted}
\usepackage{listings}

\usepackage{pifont}

\def\paperID{42359} 
\def\confName{CVPR}
\def\confYear{2025}

\title{Beyond Caption-Based Queries for Video Moment Retrieval}

\author{David Pujol-Perich$^{\gamma,\delta}$\thanks{Work partially completed whilst at University of Bristol.} \quad Albert Clapés$^{\gamma,\delta}$ \quad Dima Damen$^{\zeta}$ \quad Sergio Escalera$^{\gamma,\delta}$ \quad Michael Wray$^{\zeta}$ \\
$^{\gamma}$University of Barcelona \quad $^{\delta}$Computer Vision Center \quad $^{\zeta}$University of Bristol\\
{\tt \small david.pujolperich@ub.edu}\\
}

\begin{document}

\maketitle

\setlength\abovedisplayskip{4pt}
\setlength\belowdisplayskip{4pt}

\input{sec/abstract}  
\input{sec/introduction}
\input{sec/related_work}

\input{sec/problem_formulation}

\input{sec/empirical_analysis_of_vrms}

\input{sec/experimentation}
\input{sec/conclusions}
\input{sec/acknowledgements}

{
    \small
    \bibliographystyle{ieeenat_fullname}
    \bibliography{main}
}

\clearpage
\noindent {\Large \textbf{Supplementary Material}\par}
\vspace{0.2cm}
\appendix

\renewcommand*{\thesection}{\Alph{section}}
\renewcommand*{\thefigure}{\Alph{figure}}
\renewcommand*{\thetable}{\Alph{table}}
\setcounter{section}{0}
\setcounter{table}{0}
\setcounter{figure}{0}

\input{sec/final_supp/intro_supp}
\input{sec/final_supp/expanded_descriptions_metrics}
\input{sec/final_supp/qualitative_results_VMR}
\input{sec/final_supp/implementation_details}
\input{sec/final_supp/expanded_main_results}
\input{sec/final_supp/ablation_realism_of_queries}
\input{sec/final_supp/impact_of_calibration_in_collapse}
\input{sec/final_supp/extended_results_dinsentangling_single_multi}

\input{sec/final_supp/extended_language_vs_multiplicity_gap}
\input{sec/final_supp/extended_ablation_effect_of_n_decoder_queries}
\input{sec/final_supp/qualitative_results_search_queries}

\end{document}

%% file: sec/abstract.tex
\begin{abstract}
Current Video Moment Retrieval (VMR) models are trained on videos paired with captions, which are written by annotators after watching the videos. These captions are used as textual queries---which we term \textbf{caption-based queries}. This annotation process induces a visual bias, leading to overly descriptive and fine-grained queries, which significantly differ from the more general \textbf{search queries} that users are likely to employ in practice. 

In this work, we investigate the degradation of existing VMR methods, particularly of DETR architectures, when trained on caption-based queries but evaluated on search queries. For this, we introduce three benchmarks by modifying the textual queries in three public VMR datasets---i.e., HD-EPIC, YouCook2 and ActivityNet-Captions.
Our analysis reveals two key generalization challenges: (i) A language gap, arising from the linguistic under-specification of search queries, and (ii) a multi-moment gap, caused by the shift from single-moment to multi-moment queries. We also identify a critical issue in these architectures---an active decoder-query collapse---as a primary cause of the poor generalization to multi-moment instances. We mitigate this issue with architectural modifications that effectively increase the number of active decoder queries. Extensive experiments demonstrate that our approach improves performance on search queries by up to $14.82\%$ $mAP_m$, and up to $21.83\%$ $mAP_m$ on multi-moment search queries. The code, models and data are available in the \href{https://davidpujol.github.io/beyond-vmr/}{project webpage}.

\end{abstract}

%% file: sec/introduction.tex
\vspace{-0.6cm}
\section{Introduction}
\begin{figure}[t]
\centering
\includegraphics[width=0.48\textwidth]{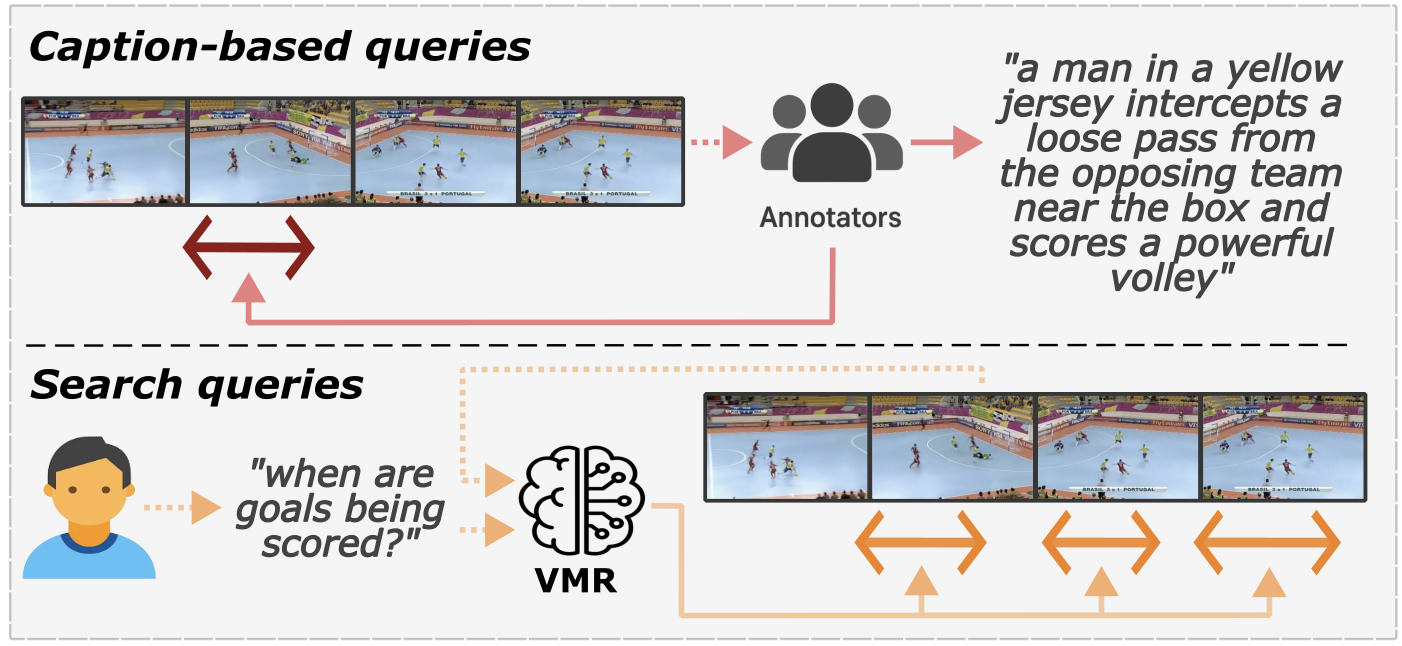}
\vspace{-0.7cm}
\caption{\label{fig:main_fig} After watching a video, annotators write detailed, visually-informed captions that map to a single GT moment. However, at inference time, users formulate less detailed, visually-uninformed search queries that often map to multiple GT moments.}
\vspace{-0.6cm}
\end{figure}

Video Moment Retrieval (VMR) aims to localize temporal segments in a video given a user-defined textual query. While current models achieve remarkable success on existing benchmarks, in this work we raise awareness of a key limitation of VMR datasets: text queries are defined using the annotated captions, written after annotators watch the videos. These captions, which we name \textit{caption-based queries}, induce a visual bias---overly descriptive visually-informed textual annotations. In contrast, real users interact through \textit{search queries}, often formulated without watching the video, which can be of a more general and \emph{under-specified} nature~\cite{sajjad2012underspecified}. For instance, while an annotator might formulate a caption-based query \textit{``a man in a yellow jersey intercepts a loose pass from the opposing team near the box and scores a powerful volley''}, a typical search query can be \textit{``when are goals being scored?''} (see Fig.~\ref{fig:main_fig}). 

Devising VMR methods that robustly perform on search queries requires suitable benchmarks, different from the existing caption-based ones~\cite{lei2021detecting, krishna2017dense,ozkose2024automatic, perrett2025hd}. However, collecting new search-query datasets remains an open challenge, as it is unclear how text annotation and video observation can be decoupled in a feasible manner. We instead re-purpose existing datasets---which include paired videos and captions---by proposing a pipeline that under-specifies captions.
We systematically explore levels of under-specificity, by changing the level of details available in the original caption.
We consequently propose three \{\textit{S}\}earch-query benchmarks: \textit{HD-EPIC-S\{1,2,3\}}, \textit{YC2-S} and \textit{ANC-S} based on HD-EPIC, YouCook2 and ActivityNet-Captions.

DETR~\cite{carion2020end}
has become the cornerstone of most existing VMR methods~\cite{moon2023correlation, zhao2025ld, sun2024tr, kim2023self}, thanks to its usage of $K$ learnable decoder queries, each of which maps to a potential retrieved moment with a corresponding confidence score. Our evaluation indicates that these methods, trained on caption-based queries, substantially degrade when evaluated on more under-specified search queries. We identify two key factors driving this degradation: (i) a \textit{language gap}, reflecting the linguistic distribution shift between caption and search queries, and (ii) a \textit{multi-moment gap}, arising from how caption-based queries map to a single ground-truth (GT) moment, while the under-specified search queries often map to multiple moments.  

In this paper, we quantify the impact of both the language and multi-moment gaps, and particularly address the former by proposing architecture modifications including the removal of self-attention mechanisms and the addition of a decoder-query dropout regularizer. These modifications improve generalization to search queries without the expensive task of re-annotating VMR training sets. 

In short, our contributions are: 
1) we explore the task of VMR beyond using captions as textual queries. We re-formulate these as under-specified versions of existing captions, so they are closer to common user-defined queries, while still making the most of available benchmarks;
2) we create three VMR benchmarks with search queries by mapping caption-based queries to under-specified search queries; 3) we demonstrate the significant degradation in performance and identify its two main causes: a \textit{language} and a \textit{multi-moment gap}; and 4) we mitigate this degradation, particularly that induced by the multi-moment gap, by introducing targeted architectural modifications that boost generalization to search queries.

%% file: sec/related_work.tex
\section{Related work}

\textbf{Video Moment Retrieval} has become a cornerstone in video understanding, aiming to localize start-end times of moments in a video, based on textual queries. Existing approaches can be broadly categorized into \textit{proposal-based} and \textit{proposal-free} methods. \textit{Proposal-based} approaches generate candidate temporal segments through temporal anchors~\cite{gao2017tall, buch2017sst, xu2019multilevel} or sliding-windows~\cite{anne2017localizing, chen2018temporally}. However, their performance heavily depends on the quality and redundancy of these proposals. In contrast, \textit{proposal-free} methods avoid the explicit candidate generation, leveraging a single-stage architecture to predict the moment segments. Most of these works adopt DETR-based architectures~\cite{carion2020end}, which refine a fixed set of learnable decoder queries, each of which represents a candidate segment. This paradigm was introduced by \cite{lei2021detecting}, with follow-ups like \cite{moon2023query, moon2023correlation, zhao2025ld} improving various aspects such as the cross-modality modules, recursive decoding schemes, etc. Our work also focuses on the DETR architecture, as this is the foundation of the majority of existing state-of-the-art VMR methods~\cite{moon2023correlation, zhao2025ld, sun2024tr, zhao2024ms, tang2025sim}, thus maximizing the impact of our findings.

\textbf{Generalization of VMR methods:} While DETR-based VMR models have shown remarkable success in existing VMR benchmarks, their generalization capabilities beyond their training data distribution remains largely under-explored. Most existing works address this from a vision-centric perspective, analyzing aspects like action duration and temporal shifts~\cite{otani2020uncovering, zhang2022actionformer, lin2018bsn} or temporal biases~\cite{otani2020uncovering, yang2021deconfounded, li2023momentdiff}. More recent studies have begun exploring the role of language biases in generalization~\cite{linell2004written, liang2020towards}, tackling rare-word usage and grammatical mistakes~\cite{xu2025zero} and the use of unlabeled data to improve language robustness~\cite{bao2024vid}. In parallel, research from the multi-modal and linguistic community~\cite{pezzelle2023dealing, sanders2024grounding} emphasizes how under-specification and contextual variation also impact the generalization capabilities of models. Moreover, Liang~\etal~\cite{liang2025tvr} address the effect of imprecise queries by incorporating them into training for the task of ranked retrieval across corpora. Instead, we tackle the alternative challenge of multi-moment retrieval within a single video by generalizing to such queries in a zero-shot manner. We find that a fundamental bottleneck in VMR stems from the use of visually-informed captions as training queries. This induces a visual bias toward overly descriptive language, often misaligned with more abstract, visually-uninformed queries prompted by users in real-world scenarios.

\textbf{Query collapse in DETR:} A core issue that we address is the query collapse in DETR architectures, whereby only a small subset of decoder queries meaningfully contribute to the final prediction(s), while the rest remain inactive. This issue, reported in object detection~\cite{zhu2020deformable, meng2021conditional, li2022dn}, temporal action detection~\cite{kim2023self}, and 3D detection~\cite{xu2024redundant, zhu2023conquer}, is largely attributed to the sparse supervision from the one-to-one matching~\cite{carion2020end}. We observe a similar phenomenon in VMR, driven instead by the single-moment prior of existing benchmarks, which typically provide one annotated moment per query. This leads to a significant query collapse that hinders generalization to multi-moment queries. While some works introduce alternative mechanisms to provide additional supervision signals~\cite{jia2023detrs, chen2023group, fang2024feataug, wang2109anchor, li2022dn, zhao2024ms}, these mainly target accelerating convergence, proving unable to overcome this strong prior. Curating new datasets with multiple annotated moments per query could alleviate this issue~\cite{lei2021detecting, cao2025one}, but would entail costly re-annotations or directly discarding most existing datasets. This motivates our approach, which introduces architectural modifications that counter the single-moment prior, leveraging existing datasets while improving generalization to unseen multi-moment scenarios.

%% file: sec/problem_formulation.tex
\vspace{-0.2cm}
\section{Problem definition and benchmarking}\label{sec:problem_definition_and_benchmarking}

\subsection{Problem definition}

Video Moment Retrieval (VMR) is defined as follows: Given a video–query pair $(v_i, q_i)$, the task is to predict the start–end times $\{(s_j, e_j)\}_{j=1}^{M_i}$ of all temporal segments in $v_i$ corresponding to the textual query $q_i$---i.e. moments. In this work we revisit this task, and focus on the often overlooked aspect of how textual queries are defined.

Specifically, we highlight the underlying assumption of all existing VMR benchmarks where queries are created from the captioning annotations---i.e. annotators who watch the videos before writing a sentence that best describes the moment~\cite{gao2017tall, krishna2017dense, zhou2018towards, perrett2025hd}. These \textit{caption-based queries} induce a visual bias that the queries perfectly match the description of the moment, thus descriptive and fine-grained in nature. In contrast, real-world users formulating their textual queries are normally unaware of the detailed content of the video, relying instead on broader,  \textit{under-specified} descriptions. Such queries can range from moderately detailed to very general ones, differing substantially from captions. We refer to these as \textit{search queries}.

To model the distribution shift between caption and search queries, we derive $\mathcal{Q}_{search}$ from $\mathcal{Q}_{caption}$ through varying degrees of under-specification. We thus capture the shift from visually-informed to visually-uninformed textual queries. This allows us to study how VMR methods trained on caption-based queries $\mathcal{Q}_{caption}$ perform on search queries $\mathcal{Q}_{search}$, more closely aligned with common situations in which users either do not know or do not exactly remember the contents of the video.

\subsection{Benchmarks}
Common VMR benchmarks rely on caption-based queries, and thus do not include search-query annotations. Re-annotating benchmarks is a challenging task, as it is unclear how to disentangle search-query annotation from video observation in a feasible and scalable manner. Accordingly, we propose to make the most of existing datasets, introducing a pipeline that rewrites a densely-annotated caption-based dataset, to a search-query variant. Dense temporal annotations are essential here since, as queries become more under-specified, these may correspond to multiple additional moments in the video. If dense annotations are provided, the search for these new correspondences can be done automatically. This contrasts with sparsely annotated datasets~\cite{lei2021detecting}, which would require extensive manual re-annotation to cover unlabeled moments corresponding to the under-specified search query.
In particular, we use three densely-annotated public datasets to introduce our search-query pipelines. We detail our pipeline next.

\subsubsection{Search-query pipeline}\label{sec:search_query_pipeline}

\begin{figure}[t]
\centering
\includegraphics[width=0.48\textwidth]{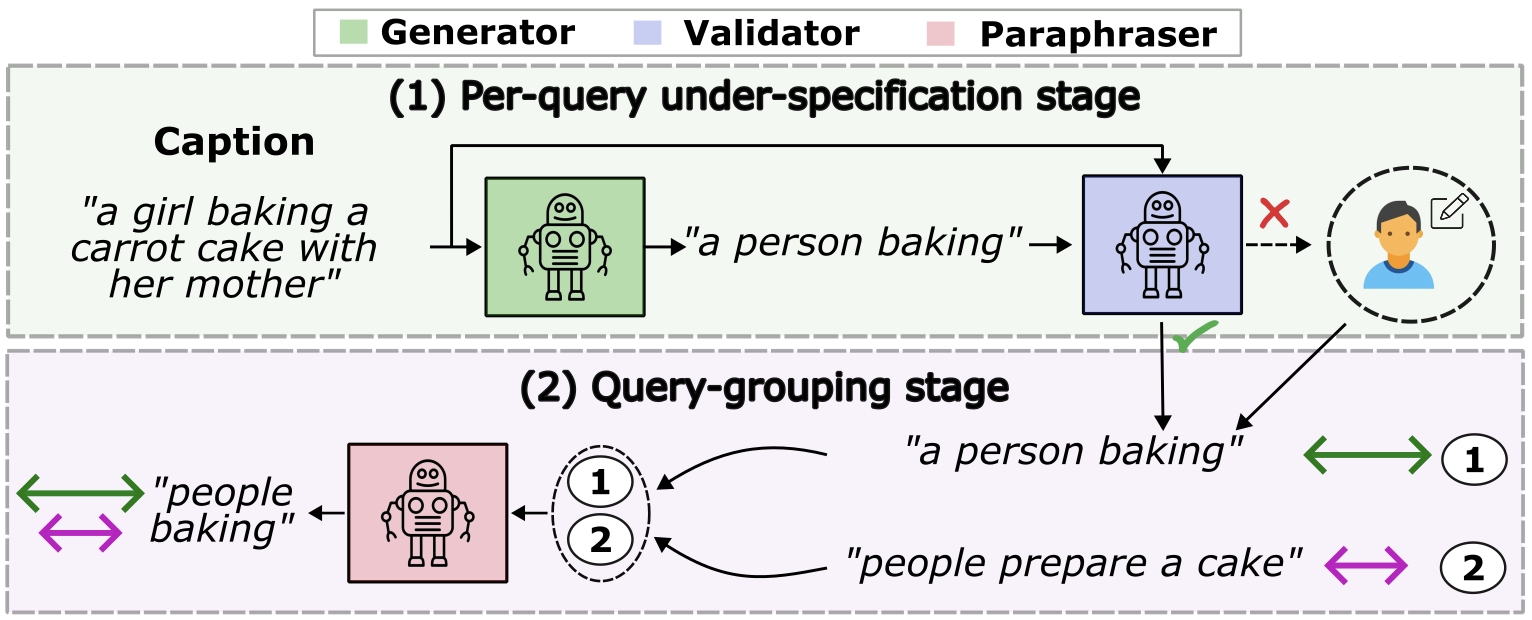}
\vspace{-0.7cm}
\caption{\label{fig:annotation_pipeline} Overview of the search-query pipeline. Each of the caption is first processed by an agent that generates per-query under-specifications, which are validated by a second identical agent and manually re-annotated if abnormal. Individual queries mapping to the same under-specified query are then grouped, and a final agent produces a representative search query per group.}
\vspace{-0.4cm}
\end{figure}

Our pipeline (see Fig.~\ref{fig:annotation_pipeline}) comprises two main stages:

\noindent 1) \textbf{Per-query under-specification stage:} To simulate the potential ambiguity of search queries, we generate under-specified queries from fine-grained ones through an LLM-based pipeline. Specifically, we instantiate two cooperative agents based on Gemma-12B~\cite{team2024gemma}: a rewriter and a validator. The rewriter agent receives a fine-grained caption and rewrites it into a less detailed version while preserving the core semantics. For example, the caption \textit{``a man tying his running shoes before starting a marathon''}, will be rewritten as \textit{``a person getting ready to exercise''}. This step allows us to model alternative under-specified search queries where users may omit context information like subject, object or intent. Inspired by \cite{cheng2024small}, we prevent hallucinations by introducing a second agent that acts as a validator. This agent flags any inconsistent rewritings, which are subsequently corrected by human annotators.

\noindent 2) \textbf{Query-grouping stage:} When a query is under-specified, it can correspond to multiple valid moments in the video, since several fine-grained situations can fit the same broader description. For example, \textit{``a person cooking food''} could match segments showing \textit{``a man sweating onions''}, or \textit{``a woman stirring soup''}. This makes it essential to group all original fine-grained queries that map to the same or highly similar under-specified query. To perform this grouping, we compute pairwise similarities across all under-specified queries using a pre-trained transformer-based sentence encoder~\cite{reimers1908sentence}. Queries with a high similarity are merged into the same group, forming a multi-moment instance. We then use an LLM-based aggregator that summarizes each group into a single representative under-specified query, removing minor differences across group members while keeping their shared semantics. Find more details, including prompts, in Sec.~\ref{supp:search_query_pipeline}.

\subsubsection{Search-based VMR benchmarks}

The proposed search-query pipeline enables us to introduce three search-query benchmarks, denoted by ``-S'':

\noindent \textbf{HD-EPIC-S\{1,2,3\}:} HD-EPIC~\cite{perrett2025hd} is a large-scale egocentric dataset featuring long cooking videos. Due to the exceptional level of detail of its annotated captions ($16.47$ words per query on avg.), we derive three progressively under-specified variants---i.e., S1, S2, S3---by gradually removing contextual details. For example, \textit{``Pick up a tissue from inside the plate on the countertop using the right hand''} $\rightarrow$ \textit{``Pick up a tissue from the plate''} (S1) $\rightarrow$ \textit{``Pick up a tissue''} (S2) $\rightarrow$ \textit{``Pick up something''} (S3).
    
\noindent \textbf{YouCook2 (YC2)-S:} YC2~\cite{ozkose2024automatic} contains step-level narrations for instructional cooking videos. YC2-S replaces the original detailed descriptions with under-specified versions, mostly emphasizing the main actions. For example \textit{``Add salt to the pan and mix''} $\rightarrow$ \textit{``Season food''}.
    
\noindent \textbf{ActivityNet-Captions (ANC)-S:} ActivityNet-Captions~\cite{krishna2017dense} consists of third-person open-domain videos. ANC-S derives under-specified versions of its corresponding textual queries, removing fine-grained contextual details. For example, \textit{``The man in white shirt is strumming the bongo drum.''} $\rightarrow$ \textit{``A person plays an instrument''}. Unlike previous benchmarks, ANC-S presents a considerable number of multi-moment search queries with overlapping moments.

Tab.~\ref{tab:stats_datasets} depicts the main statistics of these benchmarks. Using our pipeline, we extend existing single-moment datasets into new multi-moment versions, where up to $47.57\%$ of the queries correspond to multiple moments. Due to the linguistic under-specification, the average query length is also reduced by up to $82\%$. All this while ensuring the realism and reliability of our generated search queries, validated in Sec.~\ref{sec:study_of_the_realism_of_the_queries} and Sec.~\ref{supp:qualitative_search_queries}.

\begin{table}[t]
\footnotesize
\centering
\caption{Statistics of the search-based VMR benchmarks\label{tab:stats_datasets}}
\vspace{-0.3cm}
\resizebox{0.48\textwidth}{!}{
\begin{tabular}{c|cccccccccccc}
\toprule
Dataset &
\makecell{\# videos} &
\makecell{\# queries} &
\makecell{Duration \\ per video(s)} &
\makecell{Moments \\ per query} &
\makecell{Moments\\ per query \\(multi)} &
\makecell{\# multi. \\ queries} &
\makecell{\% multi. \\queries} &
\makecell{Query\\length} \\
\midrule

\rowcolor[gray]{0.9} HD\_EPIC~\cite{perrett2025hd} & 156 & 59,454 & 954 & 1.00 & 0.00 & 0 & 0.00 & $16.47\pm7.9$  \\
HD\_EPIC-S1 & 156 & 36,819 & 954 & 1.61 & 3.64  & 8,560 & 23.25 & $6.09\pm2.5$ \\
HD\_EPIC-S2 & 156 & 31,521 & 954 & 1.88 & 3.87 & 9,717 & 30.83 & $3.73 \pm 1.2$ \\
HD\_EPIC-S3 & 156 & 10,266 & 954 & 5.79 & 11.09 & 4,873 & 47.47 & $3.03 ± 1.0$ \\

\midrule
\rowcolor[gray]{0.9} ANC~\cite{krishna2017dense} & 14950 & 71,957 & 152.8 & 1.00 & 0.00 & 0 & 0.00 & $13.16 \pm 6.1$ \\
ANC-S & 14950 & 59,138 & 152.8 & 1.21 & 2.45 & 8,818 & 14.91 & $4.83 \pm 1.4$  \\

\midrule
\rowcolor[gray]{0.9}YC2~\cite{ozkose2024automatic} & 2268 & 13,829 & 326 & 1.00 & 0.00  & 0 & 0.00 & $8.86 \pm 3.97$ \\
YC2-S & 2268 & 7,466 & 326 & 1.84 & 2.97 & 3,212 & 43.02 & $2.08 \pm 0.50$ \\

\bottomrule

\end{tabular}}

\vspace{-0.4cm}
\end{table}

\subsection{Metrics for Search-Based VMR}

VMR performance is typically measured using Recall@1 (R1) and mean Average Precision (mAP), capturing top-1 accuracy and overall ranking quality, respectively. However, these metrics are inadequate when evaluating multi-moment queries. When retrieving under-specified queries these often map to additional GT moments. For example, while caption-based queries might target \textit{``a man in a black shirt enters through the kitchen door''} retrieving a single moment, a more general search query \textit{``a person entering a room''} could naturally map to multiple moments, including the previous. This setting thus requires metrics that estimate how individual moments are retrieved, regardless of whether they arise alone or alongside other moments. 

Existing metrics like R1 or mAP are unsuitable for two reasons: First, recall metrics like R1 only evaluate accuracy over the top-$k$ predictions---i.e., $k=1$ for R1. While appropriate when queries map to exactly $k$ moments, these metrics provide an incomplete evaluation when queries map to more than $k$ moments. For instance, for a 2-moment query, R1 assesses if the top-1 prediction matches \emph{any} GT, ignoring if the other GT was retrieved at all. Second, metrics like $mAP$ aggregate all GT moments of a query into a single video-query score, obscuring per-moment retrieval quality. Consider a query $q_1$ that maps to a GT moment $g_1$. If a model fails to retrieve it, mAP would clearly indicate failure. However, for a more general query $q_2$ that maps 4 moments ($g_1$-$g_4$), the same model might detect $g_2$-$g_4$ but still miss $g_1$. In this case, mAP would remain high, masking the error of $g_1$. Thus, the retrieval quality of an individual moment depends on how many moments co-occur with it, making it unsuitable for a fair evaluation.

\noindent We overcome these issues by introducing their respective multi-moment extensions, denoted as $R_m$ and $mAP_m$:

\noindent \textbf{Multi-moment recall ($R_m$):} $R_m$ generalizes the R1 metric to handle textual queries associated with multiple GT moments. Concretely, instead of considering the entire query as correct if one of the GT moments is retrieved with the highest confidence, $R_m$ evaluates each GT moment independently, checking if at least one of the top predictions correctly matches it. More specifically, we consider a given moment $g_i$ to be correctly retrieved under a certain IOU threshold $\tau$ ($R_m(g_i, \tau) = 1)$, if 1) the prediction detecting it had the highest confidence, or 2) all the predictions with higher confidences successfully retrieved other GT moments. The latter avoids the predictions that retrieved other valid GT moments---as there may be multiple ones--- penalize the retrieval quality of $g_i$. 

Averaging across all GT moments $\mathcal{G}$ of the dataset yields:
\begin{align}
    \small
    R_m(\tau)= \frac{1}{|\mathcal{G}|} \sum_{g_i \in \mathcal{G}} R_m(g_i, \tau)
\end{align}

Intuitively, $R_m$ measures whether a model retrieves each GT moment with a top-confidence prediction, without interference from the other GT moments in the video.

\noindent \textbf{Multi-moment mAP ($mAP_m$):} To make mAP sensitive to multiple GT, $mAP_m$ evaluates each GT individually. For a given $g_i$ and threshold $\tau$: (1) predictions intersecting $g_i$ with IOU$\ge \tau$ are considered true positives, (2) predictions not matching any GT are considered false positives, and (3) predictions matching other GT moments (not $g_i$) are ignored to avoid penalizing $g_i$ for a different, yet correct, prediction. 

With this, we compute the precision-recall curve $(P_i, R_i)$ for each $g_i$, and apply the standard mAP interpolation~\cite{everingham2010pascal} to obtain its individual score $AP_m(g_i, \tau)$. We then average across these per-GT scores for a given threshold $\tau$:
\begin{align}
\small
mAP_m(\tau) = \frac{1}{|\mathcal{G}|} 
\sum_{g_i \in \mathcal{G}} AP_m(g_i, \tau),
\end{align}
Intuitively, $mAP_m$ measures how well a model retrieves each of the GT moments, without interference from the other GT moments. See Sec.~\ref{supp:expanded_description_metrics} for more details.

%% file: sec/empirical_analysis_of_vrms.tex
\section{Search-Based VMR}\label{sec:search_based_vmr}

Here, we analyze how current VMR methods trained on caption-based datasets perform when evaluated on search queries, and how to mitigate the multi-moment gap.

\subsection{Evaluating caption-based models}\label{sec:empirical_analysis_vmr}
In our experiments, we evaluate two representative models---i.e., CG-DETR~\cite{moon2023correlation} and LD-DETR~\cite{zhao2025ld}---on the three proposed benchmarks. For each, we train on the training set of the corresponding caption-based dataset and evaluate on the same test set for both the original caption-based dataset and our proposed search-query benchmark---e.g., train on the training set of HD-EPIC, and evaluate on the test set of HD-EPIC and HD-EPIC-S2. For ANC-S and YC2-S we leverage the original splits from \cite{krishna2017dense, yoon2022selective}, while for HD-EPIC we create an 80-20 train/test split.

\begin{figure}[t]
\centering
\includegraphics[width=0.48\textwidth]{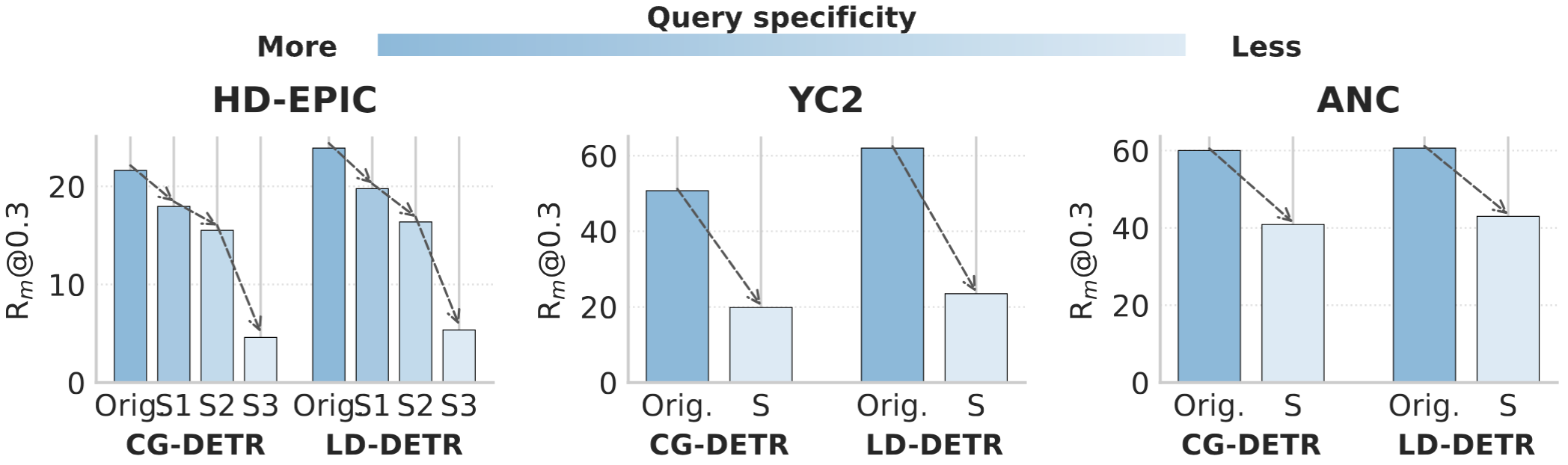}
\vspace{-0.7cm}
\caption{Evaluation of the representative models on both the original datasets and their corresponding search query extensions. \label{fig:search_query_degradation}}
\vspace{-0.3cm}
\end{figure}

Figure~\ref{fig:search_query_degradation} (left) shows the progressive performance decay across HD-EPIC-S1/S2/S3 benchmarks, with a relative degradation of up to $71.75\%$ and $77.40\%$ of $R_m@0.3$ for CG-DETR and LD-DETR, respectively. Similar trends are observed on ANC-S (center) and YC2-S (right) with drops of up to $31.8\%$ and $60.7\%$ of $R_m@0.3$, respectively. These drops evidence a substantial shift between caption-based queries and search queries, showing that existing models, when trained solely on descriptive caption-based queries, significantly degrade on less detailed search queries. This inevitably hinders the deployment of existing VMR systems in real-life scenarios (see Sec. \ref{sec:main_experiments} for the full results).

\noindent Below, we isolate two main causes of this degradation:
\begin{itemize}
    \item Language gap: The linguistic shift between caption and search queries, characterized by missing visual details, looser references or the use of more abstract nouns---e.g., \textit{``food''} instead of \textit{``green peppers''}.
    \item Multi-moment gap: The gap arising from the mismatch between training on single-moment queries and evaluating on multi-moment ones, as under-specified search queries often correspond to multiple moments.
\end{itemize}

\noindent To quantify the effect of these factors, we partition the test set of the search queries into two subsets:
$\mathcal{D}^{search}_{single}$ containing under-specified queries that map to a \underline{single} GT moment, and $\mathcal{D}^{search}_{multi}$ containing search-queries that map to \underline{multiple} GT moments.

For a fair one-to-one comparison across specificity levels, we also partition the original caption-based dataset $\mathcal{D}^{caption}$ using the same moment correspondence. Thus, 
$\mathcal{D}^{caption}_{single}$ and $\mathcal{D}^{caption}_{multi}$ contain the same moments, as $\mathcal{D}^{search}_{single}$ and $\mathcal{D}^{search}_{multi}$, respectively, but paired with their more specific captions.  
This yields our evaluation setup:
\[
(\mathcal{D}^{caption}_{single}, \mathcal{D}^{search}_{single}, ) \; \text{and} \; (\mathcal{D}^{caption}_{multi}, \mathcal{D}^{search}_{multi}, ) 
\]
allowing us to measure degradation due to purely linguistic changes (``single" split) versus the compounded effect of linguistics and multi-moment mapping (``multi" split).

\begin{figure}[t]
\centering
\includegraphics[width=0.45\textwidth]{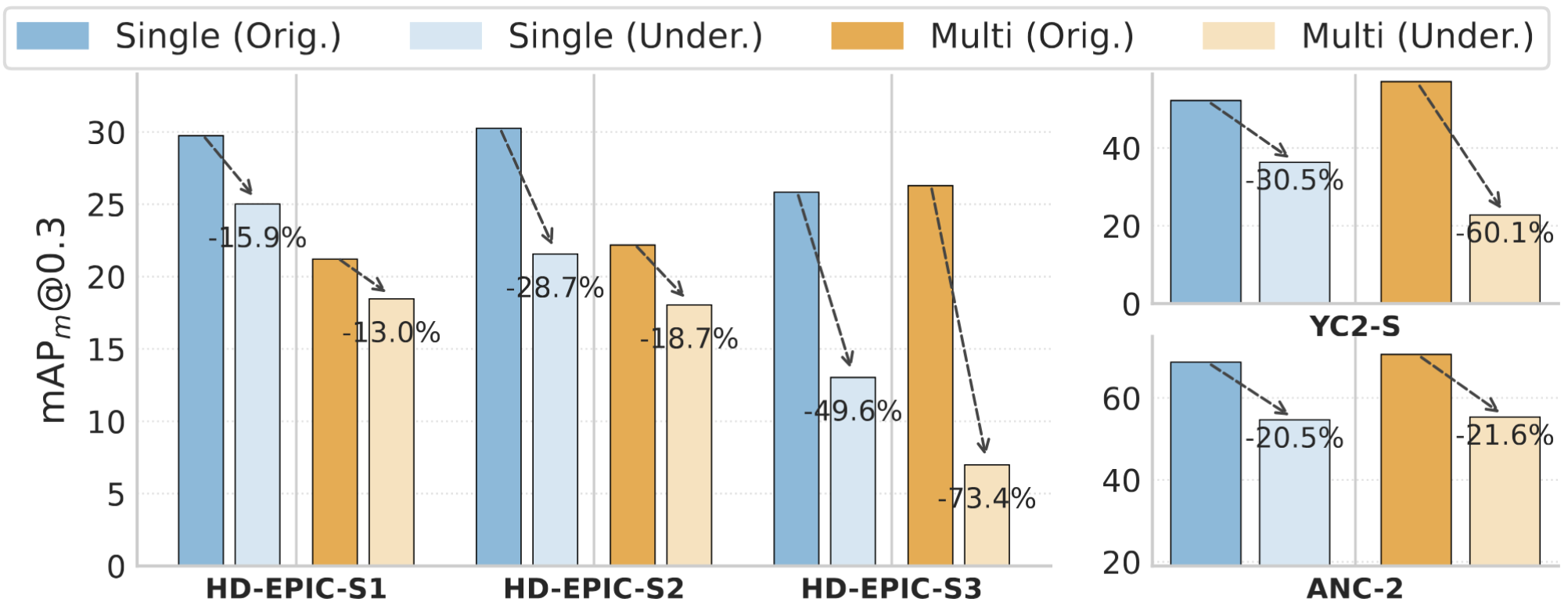}
\vspace{-0.3cm}
\caption{Performance degradation for CG-DETR on caption versus search-based evaluation for the ``single'' and ``multi'' splits. \label{fig:mult_vs_language_gap}}
\vspace{-0.4cm}
\end{figure}

Figure~\ref{fig:mult_vs_language_gap} reports the performance of CG-DETR on the three benchmarks after the decoupling of ``single'' and ``multi'' instances. For HD-EPIC, the language gap (performance on single) increases as we progressively evaluate more under-specific search queries, dropping from $15.9\%$ to $49.6\%$ with respect to the original $mAP_m@0.3$. Importantly, the compounded effect of the language and multi-moment gaps (performance on ``multi'') aggravates this effect further, reaching a degradation on HD-EPIC-S3 of $73.8\%$ compared to the original $mAP_m@0.3$. This highlights the significant additional impact on performance of the multi-moment gap. These observations remain consistent across all benchmarks. 

We next focus on addressing this multi-moment gap, an aspect largely under-explored in the VMR literature and key to the model architecture design. We leave addressing the language gap for future work, as it may be resolved with more advanced vision-language models, capable of reasoning across varying levels of specificity.

\subsection{Mitigating the multi-moment gap}

In this section, we analyze the underlying causes of the performance degradation of VMR models on multi-moment query setups. We argue that this degradation mostly stems from misalignment between caption-based training data---characterized by a single relevant moment as GT---and search-based evaluation data, which frequently contains multiple valid moments. This discrepancy induces a strong single-moment prior---i.e., a bias towards expecting a single GT moment per video-query pair. One could attempt to resolve this by curating more diverse training data, which is impractical due to annotation costs and the uncertain feasibility of devising true search-query datasets. We instead approach this issue purely from a model's perspective which allows us to reuse all existing VMR training regimes.

\subsubsection{Implications of a single-moment prior}

We next analyze why DETR-based methods trained on single-moment queries struggle with multi-moment queries at inference time. Concretely, since each decoder query produces a candidate moment, analyzing the number of active decoder queries---i.e., those whose confidence does not vanish---directly reflects the model's capacity to retrieve multiple moments. In VMR, the number of active decoder queries can also be thought as the ``compute budget'' available to retrieve all the GT moments. When the number of active queries does not scale with the number of moments of the target instance, the model cannot retrieve all moments---e.g., when only 2 queries are activated in a 4-moment instance, the upper bound of retrieved moments is $50\%$. We term this phenomenon \textit{active decoder-query collapse}.

\begin{figure}[t]
\centering
\includegraphics[width=0.38\textwidth]{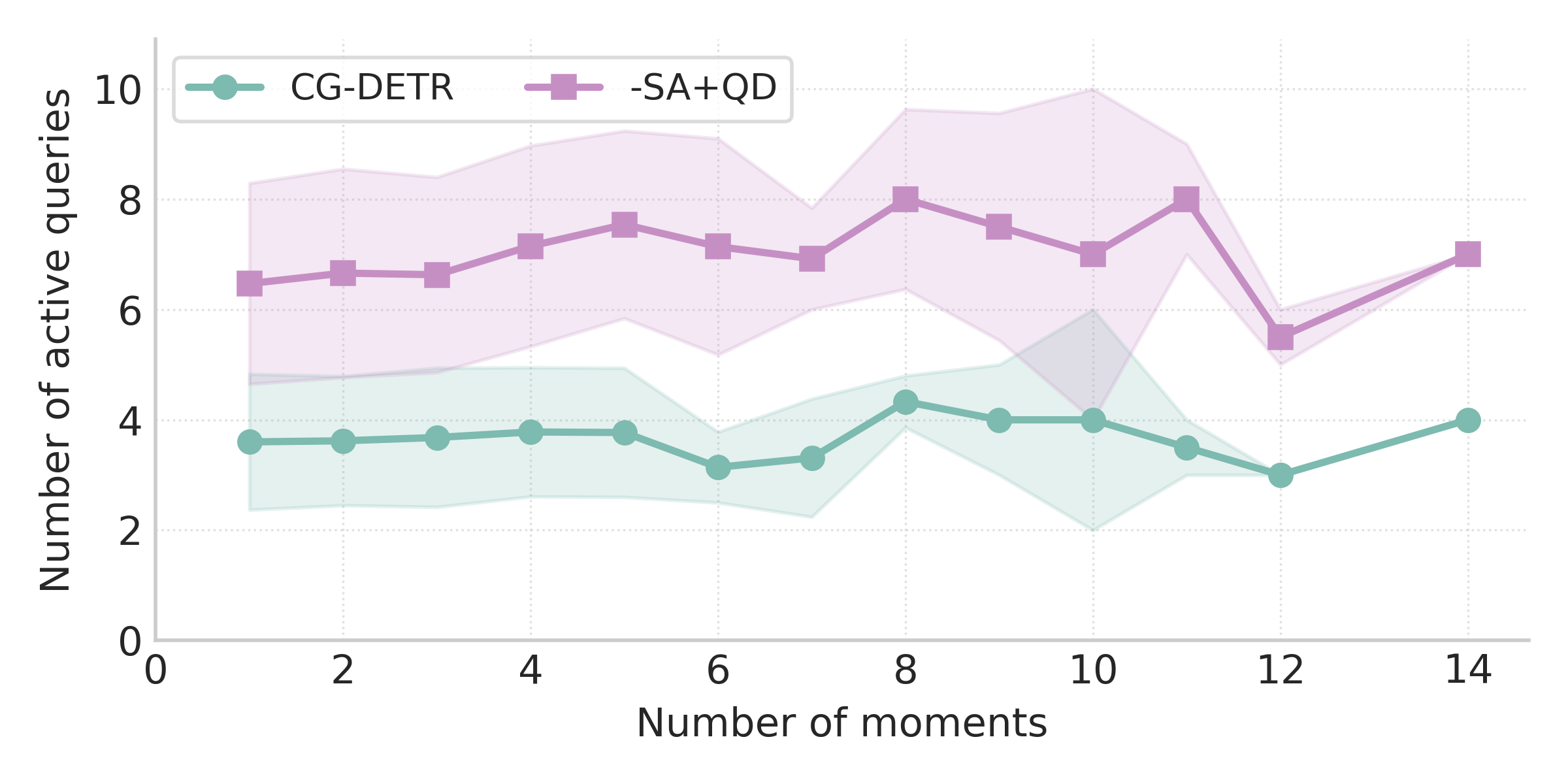}
\vspace{-0.4cm}
\caption{Visualization of the active query collapse on HD-EPIC-S2 for the base CG-DETR and our method -SA+QD. \label{fig:query_collapse}}
\vspace{-0.2cm}
\end{figure}

As shown in Fig.~\ref{fig:query_collapse}, this phenomenon indeed affects VMR methods. Concretely, VMR methods trained on standard caption-based datasets (blue line) yield an insufficient number of active decoder queries when evaluated on search queries---around 4, regardless of the number of moments of the search queries (x-axis). This stems from the single-moment training, which impedes generalization beyond queries mapping to 4 moments.

\subsubsection{Addressing the active decoder-query collapse}
Having identified the \textit{active decoder-query collapse} as a key limitation for generalization to multi-moment queries, we explore whether this can be mitigated purely from a model perspective, without altering the standard VMR training regimes and datasets. We find that this is possible by preventing models from overfitting to the single-moment prior encoded in the training data. Specifically, we identify two structural causes of this prior fitting: (i) Coordination collapse, where the self-attention mechanism causes decoder queries to suppress one another, and (ii) Index collapse, where a fixed, small subset of decoder query indices dominate activation.

In the following, we introduce architectural modifications that retain the model's capacity to learn the VMR task, while mitigating these forms of collapse.

\noindent\textbf{Coordination collapse:} The first cause arises from how the self-attention (SA) within each decoder layer enforces coordination among decoder queries. This drives them to ``agree" on which query should handle the GT moment and which should remain inactive.

Formally, a standard decoder layer can be defined as:
\begin{align}
	\small
	\hat{Q}^{l+1} = FFN(CA(SA(\hat{Q}^l), M)
\end{align}
where $M \in \mathbb{R}^{T \times F}$ are the fused multi-modal features, CA denotes cross-attention and FFN a feed-forward network.

As noted by \cite{hu2023dac}, the CA module injects the cross-modality information, while the role of SA is pushing decoder queries apart from each other to avoid redundancy. However, unintentionally, this also drives the majority of decoder queries to deactivate.

Interestingly, we find that an effective way of overcoming this issue is removing this SA module altogether, while leaving the losses unchanged:
\begin{align}
	\small
	Q^{l+1} = FFN(CA(Q^l, M)
\end{align}
Eliminating this inter-query communication prevents these coordination-based shortcuts, encouraging each decoder query to act independently. However, this also removes the model’s built-in mechanism for avoiding redundant predictions. We address this by applying Non-Maximal-Suppression (NMS) during post-processing, which filters out overlapping/redundant predictions.

\noindent\textbf{Index collapse:} Mitigating the coordination collapse alone is insufficient as the model is still able to overfit to the single-moment prior, and thus still suffers active decoder-query collapse. The reason resides in an \textit{index collapse}, where the same decoder query indices repeatedly dominate the output confidence, while the rest remain inactive---e.g., decoder queries with index 1--4 are the only ones ever activating. During training, the single-moment prior drives the model to associate the detection of the single GT moment with only a handful of fixed decoder query indices based on their learnable initializations. This dominance is progressively reinforced throughout training, leaving the rest of the decoder queries permanently inactive and thus, unused.

We counter this effect by applying a targeted query dropout strategy, which randomly zeroes out $k\%$ of the learnable queries $Q \in \mathbb{R}^{Q \times F}$ during each training iteration:
\begin{align}
\small
	\hat{Q} = Q \odot M,\quad M \sim \mathbb{B}(1-k)
\end{align}
where $\mathbb{B}$ is sampling from the Bernoulli distribution with keep probability $(1-k)$. This regularization promotes the model to distribute supervision across more queries, preventing over-reliance on a fixed subset. 

Together, these modifications considerably reduce the number of permanently inactive indices, resulting in a consistent increase in the number of active decoder queries (see Fig. \ref{fig:query_collapse} orange line), boosting search-query generalization.

%% file: sec/experimentation.tex
\section{Experimentation}\label{sec:experimentation}

\subsection{Experimental setup}

The following experiments evaluate how baselines trained on a caption-based dataset generalize to search-based benchmarks. Specifically, we evaluate HD-EPIC-S, YC2-S and ANC-S; and report the $R_m$ and $mAP_m$, on IOU $\{0.1,0.3,0.5\}$. Full implementation details are in Sec.~\ref{supp:implementation_details}.

\subsection{Main experiments}\label{sec:main_experiments}

\begin{table}[t]

\caption{Results of both CG-DETR and LD-DETR on HD-EPIC-S\{1,2,3\} benchmarks with respect to our proposed modifications.} \label{tab:main_results_hd_epic}
\vspace{-0.2cm}

\footnotesize
\centering
\label{table:comparison-results}
\resizebox{0.48\textwidth}{!}{
\begin{tabular}{c|c|c|cccc|cccc}
\toprule

\textbf{Model} & \textbf{Input} & \textbf{Variant} & \multicolumn{4}{c|}{\textbf{R$_m$}} & \multicolumn{4}{c}{\textbf{mAP$_m$}} \\
 
& & & \textbf{@0.1} &\textbf{@0.3} & \textbf{@0.5} & \textbf{Avg.} & \textbf{@0.1} & \textbf{@0.3} & \textbf{@0.5} & \textbf{Avg.} \\
\midrule

\multirow{6}{*}{CG-DETR} & \multirow{2}{*}{S1} & base & 28.61 & 17.95 & 8.99 & 18.51 & 36.21 & 22.84 & 11.59 & 23.54 \\
& & -SA+QD & \textbf{29.87} & \textbf{19.69} & \textbf{10.86} & \textbf{20.14} & \textbf{39.74} & \textbf{26.49} & \textbf{14.87} & \textbf{27.03} \\

\cmidrule{2-11}

& \multirow{2}{*}{S2} & base & 24.71 & 15.52 & 7.89 & 16.04 & 32.15 & 20.1 & 10.29 & 20.84 \\
& & -SA+QD & \textbf{26.17} & \textbf{17.00} & \textbf{9.40} & \textbf{17.52} & \textbf{35.38} & \textbf{23.39} & \textbf{13.04} & \textbf{23.93} \\

\cmidrule{2-11}
& \multirow{2}{*}{S3} & base & 9.50 & 4.61 & 2.08 & 5.39 & 16.20 & 8.01 & 3.58 & 9.26 \\
& & -SA+QD & \textbf{10.57} & \textbf{6.52} & \textbf{3.45} & \textbf{6.84} & \textbf{17.27} & \textbf{10.65} & \textbf{5.54} & \textbf{11.15} \\

\midrule
\midrule

\multirow{6}{*}{LD-DETR} & \multirow{2}{*}{S1} & base & 29.42 & 19.77 & 10.50 & 19.89 & 36.55 & 24.50 & 13.18 & 24.74 \\
& & -SA+QD & \textbf{30.18} & \textbf{20.26} & \textbf{10.83} & \textbf{20.42} & \textbf{40.5} & \textbf{27.54} & \textbf{14.94} & \textbf{27.66} \\
\cmidrule{2-11}

& \multirow{2}{*}{S2} & base & 25.23 & 16.38 & 8.46 & 16.69 & 32.42 & 21.11 & 10.93 & 21.48 \\
& & -SA+QD & \textbf{26.36} & \textbf{16.98} & \textbf{8.87} & \textbf{17.40} & \textbf{36.37} & \textbf{23.75} & \textbf{12.54} & \textbf{24.22} \\
\cmidrule{2-11}
& \multirow{2}{*}{S3} & base & \textbf{10.44} & \textbf{5.37} & \textbf{2.58} & \textbf{6.13} & 16.48 & 8.65 & 4.11 & 9.74 \\
& & -SA+QD & \textbf{10.44} & 5.28 & 2.39 & 6.03 & \textbf{17.79} & \textbf{9.06} & \textbf{4.19} & \textbf{10.34} \\

\bottomrule

\end{tabular}}

\vspace{-0.4cm}
\end{table}

\begin{table}
\footnotesize
\centering
\caption{Results of both CG-DETR and LD-DETR on YC2-S with respect to our proposed modification. \label{tab:main_results_yc2}}
\vspace{-0.2cm}
\resizebox{0.48\textwidth}{!}{
\begin{tabular}{c|c|cccc|cccc }
\toprule

\textbf{Model} & \textbf{Variant} & \multicolumn{4}{c}{\textbf{R$_{m}$}} & \multicolumn{4}{c}{\textbf{mAP$_m$}} \\
 
& & \textbf{@0.1} &\textbf{@0.3} & \textbf{@0.5} & \textbf{Avg.} & \textbf{@0.1} & \textbf{@0.3} & \textbf{@0.5} & \textbf{Avg.} \\
\midrule

\multirow{2}{*}{CG-DETR} & base & 28.92 & 19.87 & 11.22 & 20.00 & 38.83 & 26.96 & 15.21 & 27.00 \\
& -SA+QD & \textbf{29.97} & \textbf{20.32} & \textbf{11.38} & \textbf{20.55} &\textbf{41.00} & \textbf{29.40} & \textbf{17.21} & \textbf{29.20} \\

\midrule
\midrule

\multirow{2}{*}{LD-DETR} & base & 33.13 & 23.48 & 11.70 & 22.70 & 41.69 & 30.04 & 15.58 & 29.10 \\
& -SA+QD & \textbf{35.86} & \textbf{24.76} & \textbf{13.17} & \textbf{24.59} & \textbf{45.66} & \textbf{33.09} & \textbf{18.74} & \textbf{32.49} \\

\bottomrule

\end{tabular}}

\vspace{-0.2cm}
\end{table}

\begin{table}[t]
\footnotesize
\centering

\caption{Results of both CG-DETR and LD-DETR on ANC-S with respect to our proposed modification. \label{tab:main_results_anet}}
\vspace{-0.2cm}
\resizebox{0.48\textwidth}{!}{
\begin{tabular}{c|c|cccc|cccc }
\toprule

\textbf{Model} & \textbf{Variant} &  \multicolumn{4}{c}{\textbf{R$_{m}$}} & \multicolumn{4}{c}{\textbf{mAP$_m$}} \\
 
& & \textbf{@0.1} &\textbf{@0.3} & \textbf{@0.5} & \textbf{Avg.} & \textbf{@0.1} & \textbf{@0.3} & \textbf{@0.5} & \textbf{Avg.} \\
\midrule

\multirow{2}{*}{CG-DETR} & base & 60.44 & 40.89 & 24.56 & 41.96 & 72.12 & 54.92 & 36.42 & 54.48 \\ 
& -SA+QD & \textbf{63.75} & \textbf{43.12} & \textbf{25.50} & \textbf{44.12} &\textbf{74.00} & \textbf{56.42} & \textbf{38.20} & \textbf{56.20}  \\

\midrule
\midrule

\multirow{2}{*}{LD-DETR} & base & 62.58 & 43.00 & \textbf{26.08} & 43.88 & 73.35 & 56.17 & \textbf{36.79} & 55.43 \\
& -SA+QD & \textbf{65.21} & \textbf{43.89} & 25.77 & \textbf{44.95} & \textbf{74.25} & \textbf{56.31} & 36.69 & \textbf{55.75} \\

\bottomrule

\end{tabular}}

\vspace{-0.5cm}
\end{table}

\textbf{Do our proposed modifications bridge the generalization gap to search queries?}
From the results, we observe that our proposed architectural modifications, hereby denoted as (-SA+QD), substantially improve performance across all search-query datasets. For instance, on HD-EPIC-S2 (see Tab. \ref{tab:main_results_hd_epic}) these modifications increase $R_m@0.1$ from $24.71$ to $26.71$ and the $mAP_m@0.1$ from $32.15$ to $35.38$. Similarly, on YC2-S (see Tab. \ref{tab:main_results_yc2}) we observe an absolute improvement of up to $2.96$ $mAP_m@0.3$. 
Moreover, even with the smaller multi-moment gap for ANC-S, we observe that our modifications lead to comparable or improved performance across all metrics (see Tab.~\ref{tab:main_results_anet}).
To contextualize these gains, we also compare against an oracle of the base model, which is trained directly on the under-specified queries---thus perfectly matching the evaluation specificity, unlike the models trained on captions. Our approach recovers nearly $70\%$ of the oracle gap, confirming the effectiveness of (-SA+QD) in bridging the multi-moment gap, thus improving generalization to search queries. Full results can be found in Sec.~\ref{supp:expanded_main_results}.

\begin{figure}[t]
\centering
\includegraphics[width=0.48\textwidth]{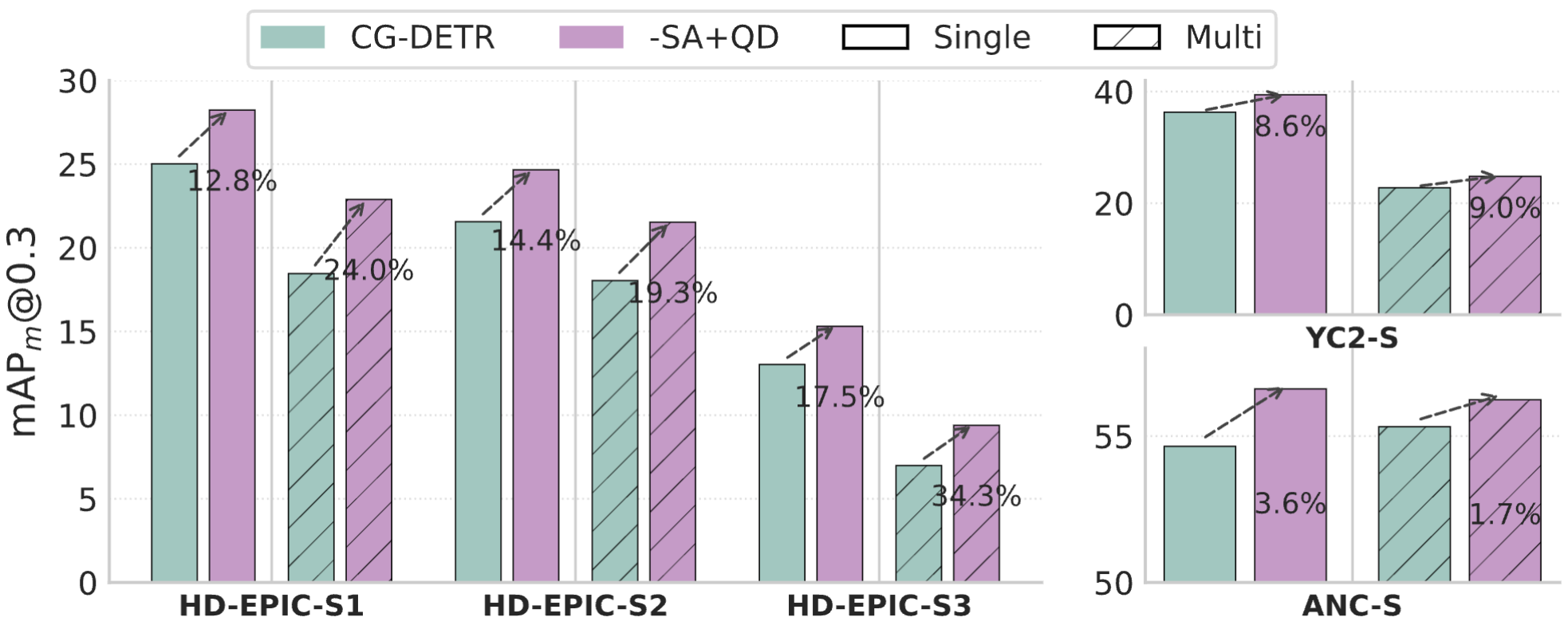}
\vspace{-0.7cm}
\caption{Dissection of the performance of CG-DETR on HD-EPIC-S2, for single, multi and all the instances, respectively. \label{fig:single_vs_multi_cg_detr_u2}}
\vspace{-0.1cm}
\end{figure}

\textbf{Where do these gains come from?}
To disentangle the benefits of our proposed modifications, we separately evaluate single and multi-moment instances. Figure~\ref{fig:single_vs_multi_cg_detr_u2} shows that while performance of (-SA+QD) on single-moment queries improves modestly, in most cases there is a prominent improvement on multi-moment queries by up to $34.3\%$ $mAP_m@0.3$. This confirms that our method specifically benefits multi-moment queries while also improving single-moment cases. See Sec.~\ref{supp:expanded_main_results}. for the extended results.

\subsection{Ablations}
Below, we ablate key aspects of our findings, reporting results for CG-DETR on HD-EPIC-S2. We report the average $R_m$ and $mAP_m$ across IoU values \{0.1, 0.3, 0.5\}. Additional ablations can be found in Supp.

\noindent \textbf{Alternative methods for decoder query activation:}
We examine whether alternative methods can increase the number of active queries and yield comparable gains to our proposal. Specifically, we evaluate two families of approaches (see Tab. \ref{tab:ablation_alternative_ways_main_text}): (i) alternative matching strategies that provide supervision to multiple decoder queries, and (ii) a data-augmentation scheme that simulates multi-moment setups. 

While matching strategies increase the number of active queries, these activated queries produce redundant predictions---predicting nearly identical segments around the same GT moment. This can be observed in the number of predictions overlapping a GT ($\% \text{match P}$), which nearly doubles, while the number of retrieved GT moments ($\% \text{match GT})$ decreases, leading to a drop in generalization.

Similarly, data-augmentation techniques that replicate GT moments in different video locations also fail to improve generalization. We attribute this to the disruption of temporal coherence, which leads to overfitting.

Overall, these results confirm that merely increasing the number of active queries through additional supervision is insufficient; effective generalization to multi-moment setups also requires diversity-promoting mechanisms that encourage complementary behavior in decoder queries.

\begin{table}[t]
\footnotesize
\centering

\caption{Ablation of methods to increase number of active queries.
\label{tab:ablation_alternative_ways_main_text}}
\vspace{-0.7cm}
\resizebox{0.48\textwidth}{!}{
\begin{tabular}{c|cc|ccc }
\toprule

\textbf{Variant} & \multicolumn{1}{c}{\bm{$R_m$}} & \multicolumn{1}{c|}{\bm{$mAP_m$}} & \textbf{\# active} & \textbf{\% match P} & \textbf{\% match GT} \\
 
\midrule

base & 16.04 & 20.84 & $3.64\pm1.18$ & $0.06\pm0.07$ & $0.36\pm0.35$ \\
\midrule

+ 1-to-5 matching~\cite{li2022dn} & 14.66 & 16.30 & $9.56\pm3.20$ & $0.11\pm0.16$ & $0.21\pm 0.28$ \\
+ 1-to-k matching~\cite{li2022dn} & 10.78 & 11.01 & $20.00\pm0.0$ & $0.16\pm0.33$ & $0.07\pm0.18$  \\
+group\_matching~\cite{chen2023group} & 15.34 & 17.97  & $8.69\pm3.08$ & $0.10\pm0.13$ & $0.27\pm 0.31$ \\
+hybrid~\cite{jia2023detrs} & 14.67 & 17.04 & $8.68\pm2.90$ & $0.10\pm0.13$ & $0.28\pm0.32$  \\
+ms\_matcher~\cite{zhao2024ms} & 15.75 & 20.94 & $3.58\pm1.14$ & $0.06\pm0.07$ & $0.36\pm 0.34$ \\
+data\_augmentation & 13.85 & 21.37 & $4.68\pm1.78$ & $0.07\pm0.08$ & $0.38\pm 0.35$ \\
\midrule
\midrule
-SA+QD (ours) & \textbf{17.52}  & \textbf{23.93}  & $6.43\pm2.16$ & $0.11\pm0.13$ & $0.42\pm0.37$ \\

\bottomrule

\end{tabular}}

\vspace{-0.2cm}
\end{table}

\noindent \textbf{Preserving diversity via 1-to-1 matching:}
The previous ablation showed that merely activating more queries does not improve generalization if diversity is not preserved. Our proposal addresses this, increasing the number of active queries while also maintaining diversity among them. This is achieved by keeping the 1-to-1 matcher~\cite{carion2020end}. This strategy enforces competition between decoder queries, preventing them from collapsing into a redundant prediction. As shown in Tab.~\ref{tab:ablation_diversity_main_text}, replacing it from our (-SA+QD) for a pure 1-to-k matcher~\cite{fang2024feataug} leads to redundant predictions, as numerous queries receive the same supervision signal. In contrast, partial relaxations that retain the 1-to-1 matching---e.g., \cite{chen2023group, jia2023detrs}---preserve competition and yield comparable results. This highlights the crucial role of the 1-to-1 matching to ensure that queries that are additionally activated by (-SA+QD) remain diverse and contribute to generalization.

\begin{table}[t]
\footnotesize
\centering

\caption{Effect of 1-to-1 matching in promoting diversity. \label{tab:ablation_diversity_main_text}} 
\vspace{-0.3cm}
\resizebox{0.48\textwidth}{!}{
\begin{tabular}{c|cc|ccc }
\toprule

\textbf{Variant} & \multicolumn{1}{c}{\bm{$R_m$}} & \multicolumn{1}{c|}{\bm{$mAP_m$}} & \textbf{\# active} & \textbf{\% match P} & \textbf{\% match GT} \\
 
\midrule

-SA+QD (ours) & 17.52 & 23.93 & $6.43\pm2.16$ & $0.11\pm0.13$ & $0.42\pm0.37$ \\

\midrule

+ 1-to-k matcher~\cite{li2022dn} & 10.38 & 10.39 & $20.00 \pm 0.00$ &  $0.14\pm0.35$ & $0.06\pm0.17$ \\
+group\_matching~\cite{chen2023group} & 17.30 & 23.71 & $12.70\pm6.13$ & $0.15\pm0.17$ &  $0.43\pm0.36$ \\
+hybrid~\cite{jia2023detrs} & \textbf{17.91} & \textbf{24.38} & $10.10\pm6.23$ & $0.14\pm0.16$ & $0.50\pm0.36$ \\

\bottomrule

\end{tabular}}

\vspace{-0.3cm}
\end{table}

\noindent \textbf{Effect of each component:}
Table~\ref{tab:ablation_components_main_text} evaluates variants that only include query-dropout (+QD) or only remove self-attention (-SA). Observe that neither component alone yields a significant performance gain as, by themselves, they do not overcome the query collapse---increasing only marginally the number of active queries. Combining both, in turn, increases $mAP_m$ by up to $3.09$ while nearly doubling the number of active queries. This confirms the need of solving both coordination and index collapse jointly.

\begin{table}[t]
\footnotesize
\centering

\begin{minipage}[t]{0.48\columnwidth}
\centering
\caption{Impact of the proposed architectural modifications.}
\vspace{-0.2cm}
\resizebox{1.15\textwidth}{!}{
\begin{tabular}{cc|cc|c}
\toprule
\textbf{-SA} & \textbf{+QD} & $\bm{R_m}$ & $\bm{mAP_m}$ & \textbf{\# active} \\
\midrule
& & 16.04 & 20.84 & 3.64$\pm$1.18 \\
\checkmark & & 15.31 & 21.02 & 3.72$\pm$1.16 \\
& \checkmark & 16.50 & 21.43 & 3.77$\pm$1.28 \\
\checkmark & \checkmark & \textbf{17.52} & \textbf{23.93} & 6.43$\pm$2.16 \\
\bottomrule
\end{tabular}}

\label{tab:ablation_components_main_text}
\end{minipage}
\hfill
\begin{minipage}[t]{0.48\columnwidth}
\centering
\caption{Effect of the QD dropout rate.}
\vspace{-0.19cm}
\label{tab:ablation_qd_dropout_rate_main_text}
\resizebox{0.65\textwidth}{!}{
\begin{tabular}{c|cc}
\toprule
$\bm{k}$ & $\bm{R_m}$ & $\bm{mAP_m}$ \\
\midrule
0.00 & 15.31 & 21.02 \\
0.25 & \textbf{17.52} & \textbf{23.93} \\
0.50 & 0.99 & 3.84 \\
\bottomrule
\end{tabular}}

\end{minipage}
\vspace{-0.6cm}
\end{table}

\noindent \textbf{Ablation on the query dropout rate:}
In Tab.~\ref{tab:ablation_qd_dropout_rate_main_text} we ablate over the effect of various query-dropout rates---i.e., $k=\{0.0, 0.25, 0.5\}$. Observe that performance peaks at $0.25$, after which performance decays. This confirms that a light stochastic regularization encourages broader query utilization, without compromising model convergence.

\begin{figure}[t]
\centering
\includegraphics[width=0.44\textwidth]{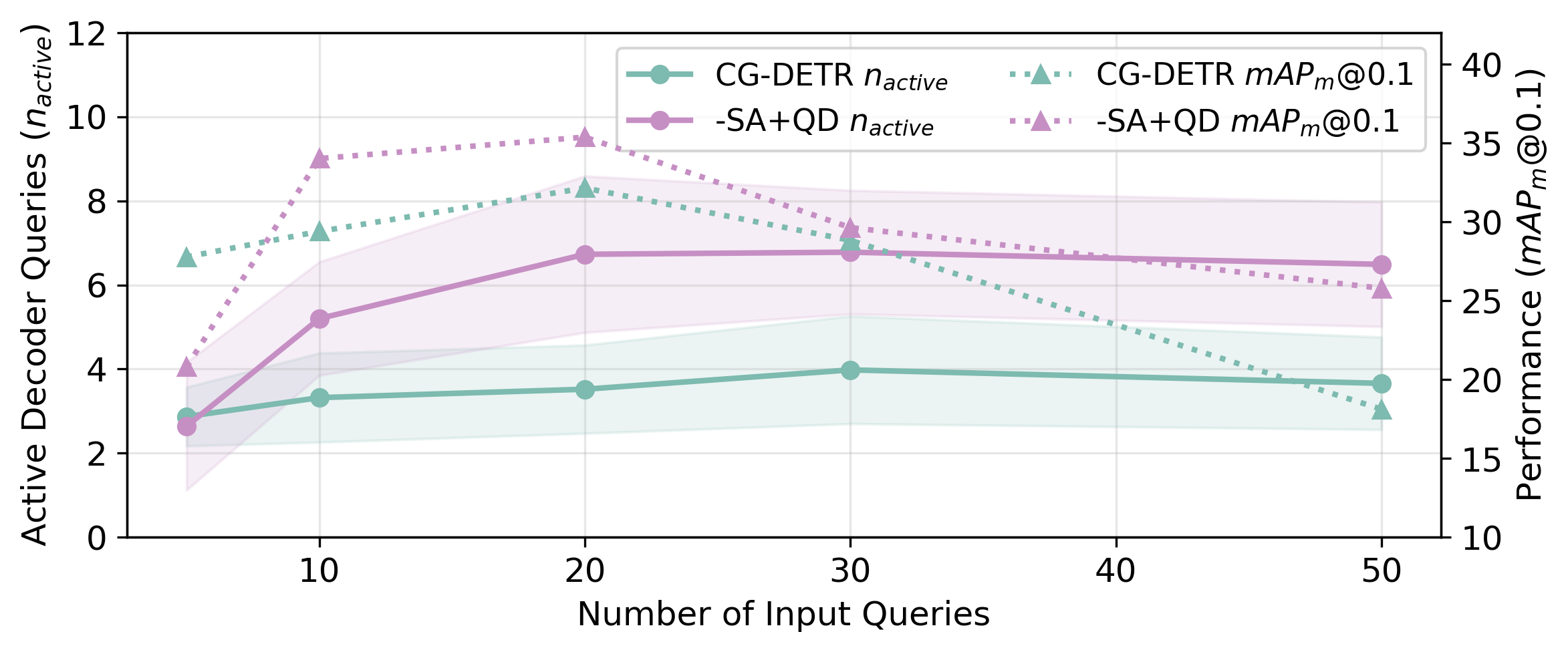}
\vspace{-0.4cm}
\caption{Evolution of the number of active queries and performance over the total number of decoder queries. \label{fig:evolution_n_active_queries}}
\vspace{-0.4cm}
\end{figure}

\noindent \textbf{Scaling the number of potential queries:}
Figure~\ref{fig:evolution_n_active_queries} investigates how the increase of the total number of decoder queries influences the number of active queries as well as performance. The base model quickly saturates. The number of active queries remains nearly constant, and performance severely degrades after peaking at $20$ queries. In contrast, our method presents a more steady increase in the number of active queries and performance ($mAP_m@0.1$). This trend holds up until $20$ queries, after which performance stabilizes. Find the extended ablation in Sec.~\ref{supp:extended_ablations}.

\begin{figure}[t]
\centering
\includegraphics[width=0.44\textwidth]{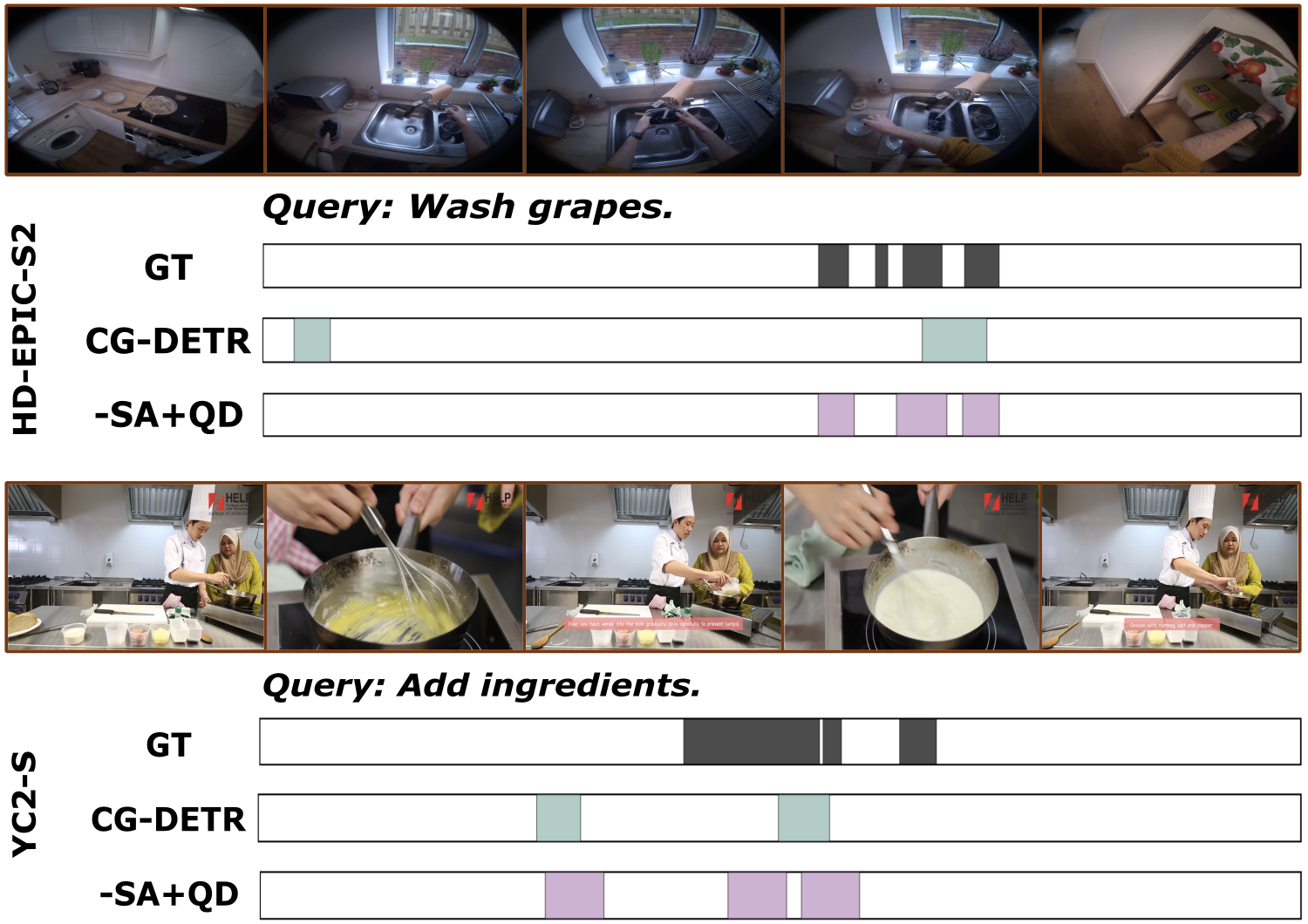}
\vspace{-0.1cm}
\caption{Qualitative results for CG-DETR on HD-EPIC-S2 and YC2-S benchmarks. \label{fig:qualitative_examples}}
\vspace{-0.5cm}
\end{figure}

\noindent \textbf{Qualitative results:} Figure~\ref{fig:qualitative_examples} presents several qualitative examples. In these, the base model---CG-DETR (CG)---shows limited success in detecting all GT moments, partially due to its small number of active queries. For example, in the second example, only two queries are activated to retrieve 3 moments. In contrast, our model activates a larger number of queries, leading to a better coverage of the GT moments. See Sec.~\ref{supp:qualitative_results_vmr} for more qualitative examples.

%% file: sec/conclusions.tex
\vspace{-5pt}
\section{Conclusions}
In this work, we revisited Video Moment Retrieval (VMR) from the perspective of generalization to search queries, moving beyond existing caption-based datasets. To this end, we introduced three search-query benchmarks, revealing a consistent performance drop when models trained on captions are evaluated on under-specified search queries. We identified and quantified two key factors for this drop: (i) a language gap between detailed and under-specified textual queries, and (ii) a multi-moment gap, arising from the shift from single-moment caption-based queries to multi-moment search queries. We further showed that the latter triggers an active decoder-query collapse. Finally, we introduced various architectural modifications that mitigated this issue, hence improving generalization to search queries---bringing VMR models closer to real-world scenarios.

%% file: sec/acknowledgements.tex
\section*{Acknowledgements}
This work has been partially supported by the Spanish project PID2022-136436NB-I00 and by ICREA under the ICREA Academia programme.
Research at Bristol is supported by EPSRC Fellowship UMPIRE (EP/T004991/1).

%% file: sec/final_supp/intro_supp.tex
\vspace{0.3cm}
This supplementary material complements the main text showing additional details and ablation studies. Concretely, Sec.~\ref{supp:expanded_description_metrics} expands on the provided description of the two introduced metrics, as well as the underlying intuition on situations where standard VMR metrics fail to provide a complete evaluation. Sec.~\ref{supp:qualitative_results_vmr} presents various additional qualitative results for each of the proposed benchmarks, and Sec.~\ref{supp:implementation_details} further implementation details regarding various aspects like the search-query pipeline, the evaluation setup, or details regarding the models and their optimization. Sec.~\ref{supp:expanded_main_results} expands on the main results presented in the main text, including the study of an oracle model trained on search queries and for comparability purposes, the performance of these models in terms of the standard mAP or the degradation study of non-DETR architectures. Sec.~\ref{sec:study_of_the_realism_of_the_queries} provides a comprehensive study of the realism and similarity of our proposed search queries with respect to real life queries. Sec.~\ref{supp:impact_calibration} studies the relationship between calibration and the observed active decoder query collapse. Sec.~\ref{supp:extended_results_disentanglement_single_multi} expands on the results presented in the main text, disentangling where the performance gains of (-SA+QD) come from by evaluating the models on single-moment, and multi-moment queries, independently. Sec.~\ref{supp:quantifying_language_multi_moment_gap} additionally provides further details on the impact of both language and multi-moment gap, while Sec.~\ref{supp:extended_ablations} similarly expands the results of all the ablation studies presented in the main text. Finally, Sec.~\ref{supp:qualitative_search_queries} conducts a qualitative study of the generated search queries for each of the proposed benchmarks.

%% file: sec/final_supp/expanded_descriptions_metrics.tex
\section{Expanded description metrics}\label{supp:expanded_description_metrics}

In this section we provide further details of the two metrics that we introduce in this work (see {Sec.~\ref{sec:problem_definition_and_benchmarking} of the main text), namely $R_m$ and $mAP_m$. Concretely, we provide additional intuition including various illustrative examples to better understand the pitfalls of existing metrics, and hence, the need of our proposed metrics. We moreover describe in further detail the formalization of both metrics.

\subsection{Intuition}

\begin{figure*}[t]
\centering
\includegraphics[width=0.7\textwidth]{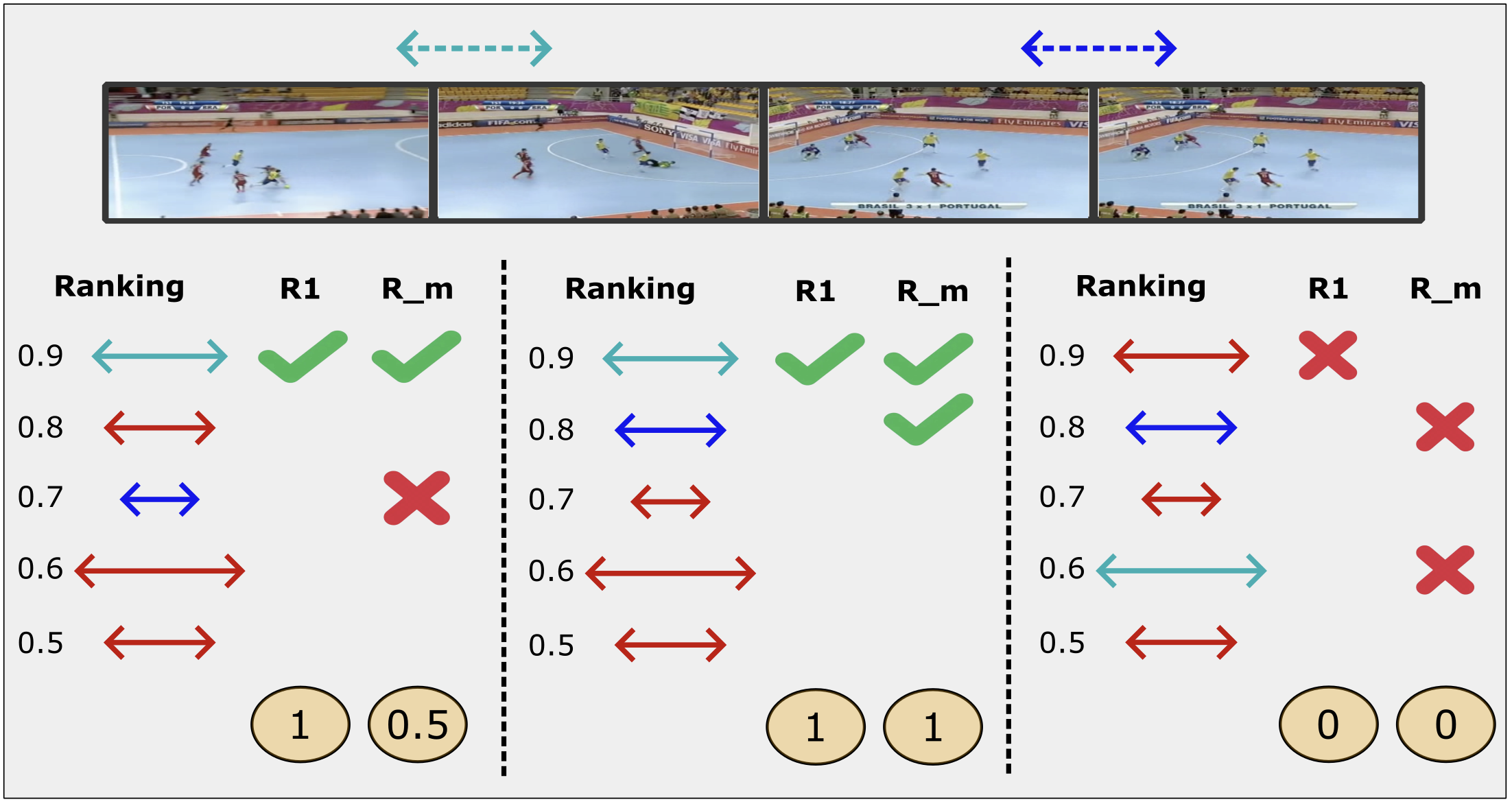}
\vspace{-0.2cm}
\caption{Intuitive examples that showcase the behavior of both $R_m$ and $R1$. Here the solid lines correspond to prediction, and the dashed ones correspond to GT moments. Moreover, in R1, the checkmark indicates that the entire query is marked as correct, while for $R_m$, since it performs a per-GT evaluation, this indicates whether the corresponding prediction ``correctly'' retrieves a GT moment or not, based on the criterion defines by the metric. The orange circles indicate the global score of the instance, which is consistent with the single score produced by $R1$, or by the average of the multiple per-GT scores for $R_m$.}
\label{fig:intuition_R_m}
\vspace{-0.2cm}
\end{figure*}

\begin{figure}[t]
\centering
\includegraphics[width=0.48\textwidth]{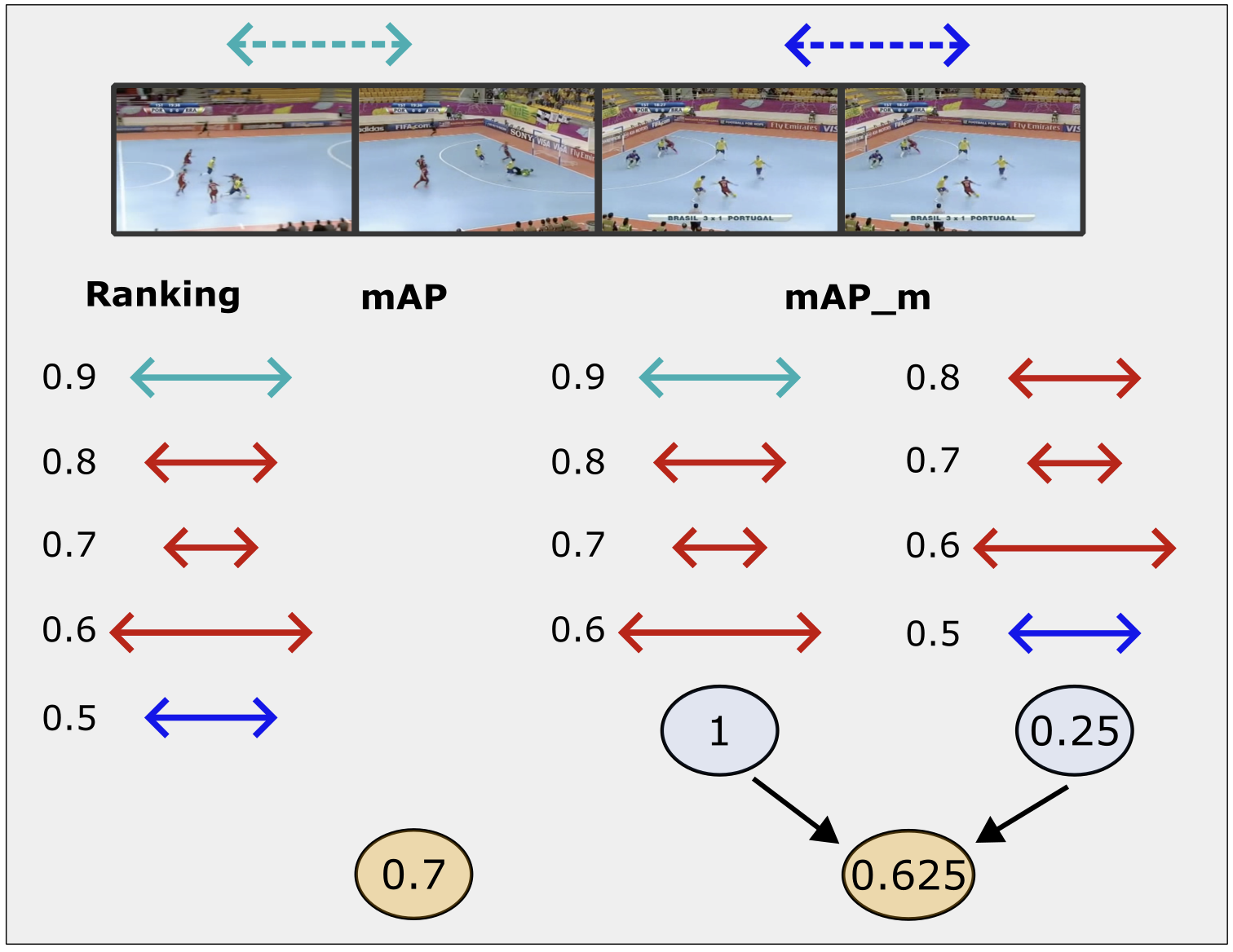}
\vspace{-0.2cm}
\caption{Example that showcases the behavior of both $mAP$ and $mAP_m$. $mAP$ computes a global score for the entire ranking, which is depicted in its corresponding orange circles. $mAP_m$, in turn, computes a score for each of the two GT moments separately. The example shows the two rankings that this evaluation leverages, effectively ignoring the influence of predictions that, while match a different GT, should not be considered invalid. The orange circle from $mAP_m$ corresponds to the average of the two respective per-GT scores. \label{fig:intuition_mAP_m}.}
\vspace{-0.2cm}
\end{figure}

Observe in Fig.~\ref{fig:intuition_R_m} an illustration of the behavior of our proposed metric $R_m$ with respect to the more standard $R1$. In the left example, observe that the highest confidence prediction matches one of the GT moments, while the second GT is matched by the third highest-prediction. In this case, standard R1 metric would predict a score of $1$, since the top prediction does correspond to one of the GT, but this score does not account for the quality of the detection of the second GT moment. $R_m$ in contrast, would provide a per-GT score, where the first moment would get a score of $1$ since it was matched to the highest-confidence prediction, while the second GT moment---matched to the third highest confidence---would get a score of $0$. The reason is that the model ranked the second prediction with a higher confidence, which corresponds to a false-positive---not matching any GT. Following the intuition of R1, this prediction has not been ``accurately'' retrieved.

The contrary happens in the second example, where $R_m$ assigns a score of $1$ for both GT, since one corresponds to the highest confidence, and the second, despite being ranked second, it is not penalized since the prediction with a higher confidence is also a match to a different GT. This hence shows similar behavior to R1.

Finally, in the third example both $R1$ and $R_m$ have a similar behavior. Since a false-positive prediction is ranked on top, all the remaining predictions that correspond to a GT get assigned a score of $0$. These examples exemplify how $R_m$ computes per-GT scores, evaluating the quality of the retrieval of each GT moment independently, and without being affected by other potential matches that may co-occur.

A similar behavior is shown in Fig.~\ref{fig:intuition_mAP_m}, which showcases the behavior of $mAP_m$ with respect to $mAP$. This considers a scenario where the model detects a given GT moment with the highest-confidence prediction, while the second is detected with the lowest one. In this case, $mAP$ computes a global score of $0.7$, where the correct prediction of one of the GT moment masks the poor detection of the other. $mAP_m$, in contrast, computes a score for each GT were the only difference is that each evaluation ignores all the predictions that match any other GT. For instance, to evaluate the dark blue moment (right most example), $mAP_m$ ignores the prediction that corresponds to the light blue one. This avoid penalizing matches that, while different, are still valid and should thus not be considered incorrect. Hence, in this case, the score for one of the moments is $1$, while the other is of $0.25$ giving a final $mAP_m$ score of $0.625$ when averaged across the two GT moments.

\subsection{Preliminaries}

Given a video-query pair, a VMR model outputs a set of $K$ predictions---i.e., candidate moments---, denoted as:
\begin{equation}
    \small
    \mathcal{P} = \{p_1, \dots, p_K\},
\end{equation}
where each prediction $p_i$ is a temporal segment predicted by the model, associated with a confidence score $c(p_i)$. These predictions are sorted in descending order:
\begin{equation}
    \small
    c(p_1) \ge c(p_2) \ge \dots \ge c(p_K).
\end{equation}
Moreover, this video-query pair maps to a set of GT moments $\mathcal{G}$:
\begin{equation}
    \mathcal{G} = \{g_1, \dots, g_n\}
\end{equation}
Given a certain IOU threshold $\tau$, we follow the existing literature~\cite{everingham2010pascal} and define a match of a prediction with a GT moment as:
\begin{equation}
    match(p_i, g_j, \tau) = 
\begin{cases}
1 & IOU(p_i, g_j) \ge \tau \\
0 & \text{otherwise}
\end{cases}
\end{equation}
For convenience, let us also define the cases where a prediction matches a GT moment that while valid, differs to the moment $g_j$ that is under evaluation:
\begin{equation}
    match\_other(p_i, g_j, \tau) = 
\begin{cases}
1 & \exists g_k \neq g_j \; st. \; IOU(p_i, g_k) \ge \tau \\
0 & \text{otherwise}
\end{cases}
\end{equation}

\subsection{Multi-moment recall $R_m$}
Standard Recall@1 (R1) assigns a score per video-query pair, this being $1$ if the highest-ranked prediction matches any of the GT moments, and $0$ otherwise. This provides only a partial performance overview when evaluating multi-moment queries---those matching to multiple GT moments---as this metric does provide information on whether the model was able to successfully detect all the GT moments.

The goal of the $R_m$ metric is to instead evaluate the detection quality for each of the GT moments, independently. Importantly, our metric avoids the interference of other co-occurring moments in the score assigned to the evaluation of a given GT moment.

More specifically, $R_m$ considers a given GT moment $g_j$ as correctly retrieved if it appears before any false positive predictions, ignoring predictions that do not match $g_j$ as they cannot be considered mistakes, since they match different, equally valid, GT moments.

Formally, let us define the index of the first prediction matching $g_j$:
\begin{equation}
    i^* = \min(i \;|\; match(p_i, g_j, \tau) = 1).
\end{equation}
Moreover, the index of the first false-positive is defined as:
\begin{multline}
i_j^{FP} = min(i | match(p_i, g_j, \tau) = 0 \; \wedge \\ \; match\_other(p_i, g_j, \tau) = 1).
\end{multline}

With this, the recall score for a given GT moment $g_j$ is defined as follows:
\begin{equation}
    R_m(g_j, \tau) = 
\begin{cases}
1 & i^* \ge i_j^{FP} \\
0 & \text{otherwise}
\end{cases}
\end{equation}
Finally, in order to obtain global scores for our dataset, we additionally compute an aggregated score:
\begin{equation}
    R_m(\tau) = \frac{1}{|\mathcal{G}|} \sum_{g_j \in \mathcal{G}} R_m(g_j, \tau).
\end{equation}
Note that similarly to other metrics, we still obtain a dataset-level metric, however, this score assigns an equal weight to all the GT moments in the dataset, regardless of the number of moments that co-occur with it in the same video-query pair. This is key as otherwise, a multi-moment query comprising of $10$ GT moments would have the same weight as that of a single query mapping to a single moment. Fixing this issue is key to ensure a fair comparison across levels of specificity, as well as in general, to provide a more fine-grained evaluation that looks at performance from a per-GT perspective, instead of a query-level one.

\subsection{Multi-moment mAP ($mAP_m$)}
Similarly to R1, mAP has the fundamental limitation that it also produces a query-video level score. Hence, it obscures the performance of the potentially multiple GT moments that may correspond to such query, as the good detection of a given moment can mask poor detections or even GT moments that were not detected at all. As argued in Sec. 3, this breaks comparability in our setup, even though we argue that this limitation also extends to the evaluation of multi-moment queries in general.

To overcome this issue, similarly to $R_m$ we propose evaluating the detection performance of each of the GT moments, independently, ensuring that a good/bad detection on one GT moment does not interfere with the scores of any other co-occurring moments. 

Accordingly, for a given GT moment $g_j$, we define the set of true positives $TP$ predictions as:
\begin{equation}
    TP_j = \{p_i \;|\; match(p_i, g_j, \tau) = 1\}.
\end{equation}
The false positives, being the predictions that do not match \underline{any} GT moment is defined as:
\begin{multline}
    FP_j = \{p_i \;|\; match(p_i, g_j, \tau)) = 0 \; \wedge \\ \; match\_other(p_i, g_j, \tau) = 0\}.
\end{multline}
And finally, we define the set of predictions that are ignored since, even though they do not match the moment that is currently evaluated ($g_j$), they nevertheless match another valid moment (not $g_j$). Ignoring them prevents these predictions from penalizing the metric for $g_j$, as this should neither benefit this metric, nor penalize it as a mistake, when it is a perfectly valid prediction. Formally,
\begin{equation}
    IGN_j = \{p_i \;|\; match\_other(p_i, g_j, \tau) = 1 \}
\end{equation}
With this, for each of the GT moments $g_j$ we compute the corresponding precision $P_j$ and recall $R_j$:
\begin{equation}
P_j(k) = \frac{\text{\#TP up to rank }k}{\text{\#TP up to rank }k + \text{\#FP up to rank }k}
\end{equation}
\begin{equation}
R_j(k) = \frac{\text{\#TP up to rank }k}{1} ,
\end{equation}
and following the original work \cite{everingham2010pascal}, compute the corresponding area-under-the-curve (AUC):
\begin{equation}
    AP_m(g_j, \tau) = \text{AUC}(P_j, R_j)
\end{equation}
Similarly to $R_m$, we also compute a dataset level score as:
\begin{equation}
    mAP_m(\tau) = \frac{1}{|\mathcal{G}|} \sum_{g_j \in \mathcal{G}} AP_m(g_j).
\end{equation}
This again results in a score where each of the GT moments in $\mathcal{G}$ have an equal weight, making comparisons across levels of specificity fair, as the score of the detection quality for a given GT moment $g_j$ is independent to the moments that it co-occurs with.

%% file: sec/final_supp/qualitative_results_VMR.tex
\section{Qualitative results VMR}\label{supp:qualitative_results_vmr}

\begin{figure}[t]
\centering
\includegraphics[width=0.49\textwidth]{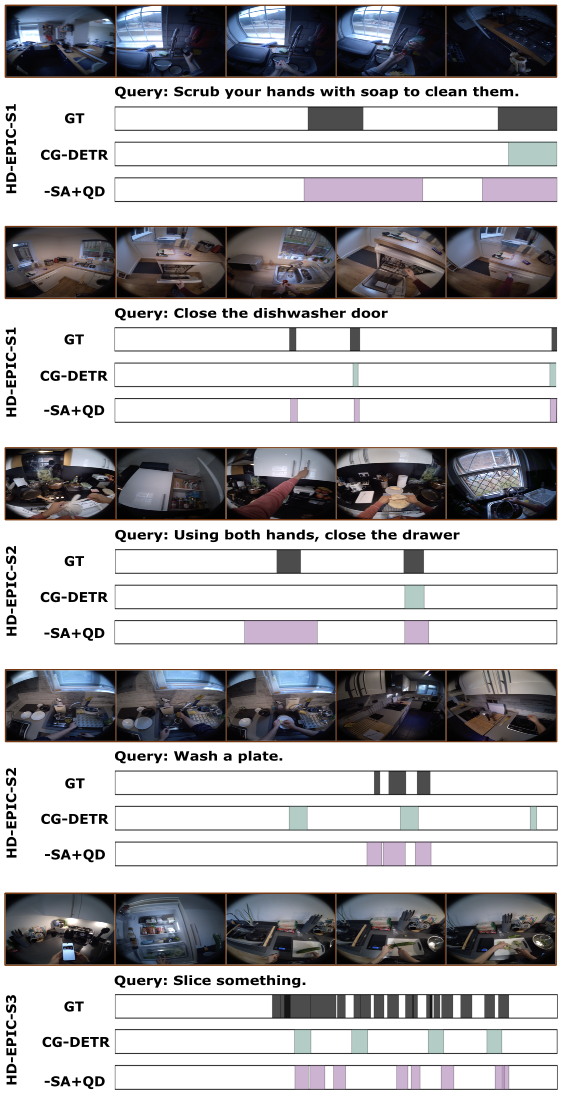}
\caption{Qualitative results of the performance of CG-DETR and our proposed modification (-SA+QD) on HD-EPIC-S1/S2/S3.\label{fig:qualitative_results_vmr1}}
\vspace{-0.2cm}
\end{figure}

\begin{figure}[t]
\centering
\includegraphics[width=0.49\textwidth]{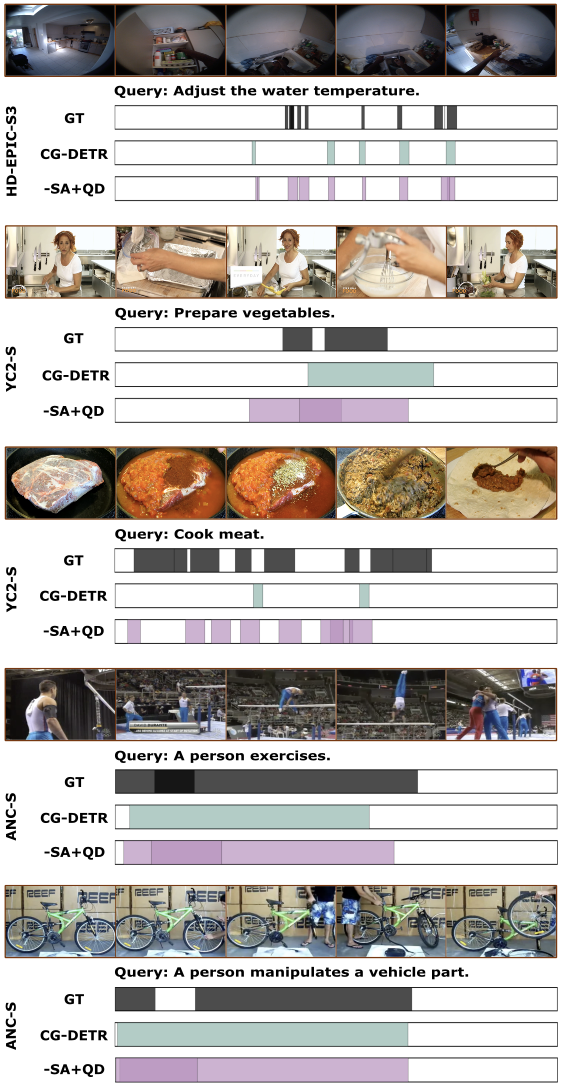}
\caption{Qualitative results of the performance of CG-DETR and our proposed modification (-SA+QD) on HD-EPIC-S3, YC2-S and ANC-S.\label{fig:qualitative_results_vmr2}}
\vspace{-0.2cm}
\end{figure}

Find below various qualitative examples that showcase the performance of our proposed modification (-SA+QD) with respect to its corresponding baseline, CG-DETR. Concretely, Fig.~\ref{fig:qualitative_results_vmr1} and Fig.~\ref{fig:qualitative_results_vmr2} show two different examples for each of the scenarios included in our proposed benchmarks---i.e., HD-EPIC-S1/S2/S3, YC2-S and ANC-S. Observe that in numerous examples, the base CG-DETR is unable to activate sufficient predictions with a non-vanishing confidence, which hinders the capacity to detect multi-moment queries as the number of active queries is smaller than the number of GT moments to retrieve. This is considerably mitigated by (-SA+QD) that consistently activates more decoder queries, thus showing considerably better behavior in these scenarios.

%% file: sec/final_supp/implementation_details.tex
\section{Implementation details}\label{supp:implementation_details}
In this section we provide additional details required for reproducibility. These include the search-query pipeline, the three proposed search-query benchmarks, and the implementation and optimization details of the evaluated baselines.

\subsection{Search-query pipeline}\label{supp:search_query_pipeline}
\subsubsection{Under-specification stage}
\textbf{Rewriter:} As described in the main text, we obtain the under-specifications of the original caption-based queries using an LLM agent based on Gemma3-12b-it. We choose this model due to its open-source availability and its good performance for this task compared to other LLMs we evaluated.

The rewriter follows an in-context learning strategy as a way to guide the agent towards the desired levels of specificity. Concretely, for each of the benchmarks we pass a small set of examples that are manually annotated by a human annotator. It is key that these examples are benchmark-specific so as to avoid the shift between the distribution of the in-context examples and that of the instances that we aim to under-specify. In Fig.~\ref{fig:prompt_underspecificity_hd_epic_s1}, Fig.~\ref{fig:prompt_underspecificity_hd_epic_s2},  Fig.~\ref{fig:prompt_underspecificity_hd_epic_s3},  Fig.~\ref{fig:prompt_underspecificity_yc2_s} and  Fig.~\ref{fig:prompt_underspecificity_anc_s} we include the prompts used for each of the benchmarks:\\

\begin{figure}[h]
\centering
\begin{lstlisting}
Simplify this query by removing or generalizing unnecessary information, but keeping the meaning of the sentence.  
Format your output as:
Simplified query: <your simplified query>

Example 1:
Query: Throw the big orange into the green recyclable bag using the left hand. With the right hand, pick up the end of the green roll and pull the roll outwards to pull one bag out of the roll using the right hand.
Simplified query: Throw the orange into the bag. Then pick up the roll and use it to pull a bag out.

Example 2:
Query: Open the upper cupboard by holding the handle of the cupboard with the left hand.
Simplified query: Open the cupboard.

Example 3:
Query: Switch the button of the socket using the left hand. This enables the power to access the food processor so as to start it.
Simplified query: Switch the button to start the food processor. 

Example 4: 
Query: Turn the dial of the food processor by turning it clockwise to switch it on. The juicer will now start rotating
Simplied query: Turn the dial by turning it to switch it on.

Example 5:
Query: Using the left hand, remove the plastic cover of the blue scissors. This action occurs at the periphery, quite off-screen.
Simplified query: Remove the cover of the scissors.

Now, simplify this query:
Query: {query}
Simplified query:
\end{lstlisting}
\caption{Rewriter prompt for HD-EPIC-S1.\label{fig:prompt_underspecificity_hd_epic_s1}}
\end{figure}

\begin{figure}[h]
\centering
\begin{lstlisting}
Simplify this query by removing or generalizing unnecessary information, but keeping the meaning of the sentence. Avoid using very vague words like ``something'' or ``items''. 
Format your output as:
Simplified query: <your simplified query>

Example 1:
Query: Throw the big orange into the green recyclable bag using the left hand. With the right hand, pick up the end of the green roll and pull the roll outwards to pull one bag out of the roll using the right hand.
Simplified query: Throw an orange away.

Example 2:
Query: Open the upper cupboard by holding the handle of the cupboard with the left hand.
Simplified query: Open a cupboard.

Example 3:
Query: Switch the button of the socket using the left hand. This enables the power to access the food processor so as to start it.
Simplified query: Switch a button on. 

Example 4: 
Query: Turn the dial of the food processor by turning it clockwise to switch it on. The juicer will now start rotating
Simplied query: Turn a dial on.

Example 5:
Query: Using the left hand, remove the plastic cover of the blue scissors. This action occurs at the periphery, quite off-screen.
Simplified query: Remove a cover.

Now, simplify this query:
Query: {query}
Simplified query:
\end{lstlisting}
\caption{Rewriter prompt for HD-EPIC-S2.\label{fig:prompt_underspecificity_hd_epic_s2}}
\end{figure}

\begin{figure}[h]
\centering
\begin{lstlisting}
Simplify this query by removing or generalizing unnecessary information, but keeping the meaning of the sentence.
Format your output as:
Simplified query: <your simplified query>

Example 1:
Query: Throw the big orange into the green recyclable bag using the left hand. With the right hand, pick up the end of the green roll and pull the roll outwards to pull one bag out of the roll using the right hand.
Simplified query: Throw items away.

Example 2:
Query: Open the upper cupboard by holding the handle of the cupboard with the left hand.
Simplified query: Open an item.

Example 3:
Query: Switch the button of the socket using the left hand. This enables the power to access the food processor so as to start it.
Simplified query: Switch something on. 

Example 4: 
Query: Turn the dial of the food processor by turning it clockwise to switch it on. The juicer will now start rotating
Simplied query: Turn something on.

Example 5:
Query: Using the left hand, remove the plastic cover of the blue scissors. This action occurs at the periphery, quite off-screen.
Simplified query: Remove an item.

Now, simplify this query:
Query: {query}
Simplified query:
\end{lstlisting}
\caption{Rewriter prompt for HD-EPIC-S3.\label{fig:prompt_underspecificity_hd_epic_s3}}
\end{figure}

\begin{figure}[h]
\centering
\begin{lstlisting}
Simplify this query by removing or generalizing unnecessary information, but keeping the core meaning of the sentence.
Format your output as:
Simplified query: <your simplified query>

Example 1:
Query: pick the ends off the verdalago
Simplified query: Add ingredient.

Example 2:
Query: pour the dressing over the salad and mix
Simplified query: Make a salat

Example 3:
Query: chop lettuce and place it in a bowl.
Simplified query: Prepare vegetables.

Now, simplify this query:
Query: {query}
Simplified query:
\end{lstlisting}
\caption{Rewriter prompt for YC2-S.\label{fig:prompt_underspecificity_yc2_s}}
\end{figure}

\begin{figure}[h]
\centering
\begin{lstlisting}
Simplify this query by removing or generalizing unnecessary information, so that the simplified query can match other overspecific ones if possible.
Format your output as:
Simplified query: <your simplified query>

Example 1:
Query: He then bends down and grabs a ball.
Simplified query: A person manipulates an object.

Example 2: 
Query: Then one man stands in a field holding a wooden object and begins twisting it.
Simplified query: A person manipulates an object.

Example 3:
Query: There was a penalty and one players attempts to hit the ball into the goal from the side.
Simplified query: People play a game.

Example 4:
Query: A group of people holding paintball guns and dressed in costume run into a staged setting as if in combat.
Simplified query: People play a game.

Now, simplify this query:
Query: {query}
Simplified query:
\end{lstlisting}
\caption{Rewriter prompt for ANC-S.\label{fig:prompt_underspecificity_anc_s}}
\end{figure}

\noindent \textbf{Validator:} To ensure the quality of the rewritings, each under-specified query together with its corresponding caption-based query is passed through a validator. This validator, based on an identical Gemma3-12b-it LLM agent, determines if the under-specified query is consistent with the caption. Valid instances are subject to a random validation process by human annotators, while the invalid instances are all manually reviewed and rewritten by a human annotator, if necessary. Note that thanks to the curated in-context examples, the number of invalid instances are few dozens per benchmark, representing less than $1\%$ of the queries. Find in Fig.~\ref{fig:prompt_validator} the prompt used for the validator.

\begin{figure}[h]
\centering
\begin{lstlisting}
Determine whether this simplified query is a valid underspecification of the original query or it is a hallucination that describes something different.  
Format your output as:
Validity: <valid | invalid>

Example 1:
Original: Throw the big orange into the green recyclable bag using the left hand.
Simplified: Throw the orange into the bag.
Validity: valid

Example 2:
Original: Switch the button of the socket using the left hand.
Simplified: A dog jumping a fence.
Validity: invalid

Now, evaluate this pair:
Original: {original}
Simplified: {simplified}

Validity:
\end{lstlisting}
\caption{Instruction of the Validator.\label{fig:prompt_validator}}
\end{figure}

\subsection{Grouping stage}
The grouping stage is further divided into two different steps:

\textbf{Similarity-based grouping:} For each video, we compute the STSB-Roberta-Large sentence embeddings~\cite{reimers1908sentence} of all the queries that occur in this video. We then compute their pairwise cosine similarities and form a graph where the nodes are the queries, and where two nodes are connected if their corresponding queries present a similarity equal or greater to 0.85. Given these connections, we use a DFS algorithm to form the groups/clusters that essentially contain the same semantics. 

\textbf{Representative search query generation:} Since members of the same group/cluster can still present some minor differences in their corresponding under-specified queries, we generate a single representative search query that corresponds to all of them. For instance, consider two very similar under-specifications \textit{``hold the pan''} and \textit{``hold the pot''}. Despite their similarity, the first includes unique information, that is not present in the seconds, and vice-versa. Thus, to create a single search query that corresponds to all the members of the group---i.e., containing shared semantics only---we employ a final identical Gemma3-12b-it agent that takes all the corresponding under-specifications of a given group and computes their final search query---e.g., \textit{``hold the kitchenware''} following the previous example. See in Fig.~\ref{fig:prompt_representative_query} the prompt that we used.

\begin{figure}[h]
\centering
\begin{lstlisting}
Combine these queries into a single unified description that applies to all the queries. Avoid adding unncessary ands/ors, trying to make it as concise as possible:
Example 1:
Queries: [Hold the pan using the left hand to cook tomatoes, Hold the pot using the right hand to cook onions]

Unified description: Hold the kitchenware using a hand to cook vegatables.   
Example 2:
Queries:  [Take off the left glove. , Take off the right glove]
Unified description: Take off the glove.  
Now create a unified description for the following queries:
Queries: {ann['original_queries']}
Unified description:
\end{lstlisting}
\caption{Prompt of the agent computing the representative search query for each of the clusters.\label{fig:prompt_representative_query}}
\end{figure}

\subsection{Evaluation setup}
Below we provide additional details on the evaluation of each of the benchmarks:

\textbf{HD-EPIC-S1/S2/S3:} We extract InternVideo2~\cite{wang2024internvideo2} features at an FPS rate of 3. Because HD-EPIC contains very long videos, we find that a naive evaluation on the entire videos yields uninformative results given the extremely low baseline scores. Since long-video VMR detection is beyond the scope of this paper, even though this constitutes a promising line of research, we trim the videos into 500-frame windows, treating each of them as independent instances. We further discard the windows that do not contain any GT moment, avoiding the important issue of dealing with the assumption that every instance contains at least one GT moment~\cite{flanagan2025moment}. This is a relevant issue that falls beyond the scope of this paper, mainly because the majority of baselines, including the ones we evaluate on do not provide built-in mechanisms to deal with these situations. 

Moreover, HD-EPIC does not provide pre-defined data splits suitable for VMR. Hence, we construct a training and testing split using an 80-20 split of the per-participant original data.  

\textbf{YC2-S}: We extract InternVideo2 features at an FPS rate of 3, and use the original training and validation splits as proposed by \cite{zhou2018towards}.

\textbf{ANC-S:}  We extract InternVideo2 features at an FPS rate of $3$, and use the original training and validation splits as proposed by \cite{krishna2017dense}.

\subsection{Models and optimization}
As described in the main text, we select CG-DETR and LD-DETR as representatives of the DETR-based VMR models. These models are trained on a single NVIDIA GeForce RTX 3090. We use all the original hyper-parameters from \cite{moon2023correlation} and \cite{zhao2025ld}, respectively, across all the benchmarks. Our method (-SA+QD) only introduces one new hyperparameter, being the QD rate, selected via grid search and kept as $0.25$ across all the benchmarks and models.

%% file: sec/final_supp/expanded_main_results.tex
\section{Experimental results}\label{supp:expanded_main_results}

\subsection{Expanded main results}
In this section we expand upon the main results presented in the main text. Although, as argued in the main text, the standard $mAP$ metric does not provide a fair evaluation of our setup, we find it helpful to include these results completeness and comparability with prior work. As shown in Tab.~\ref{tab:main_results_hd_epic_supp}, Tab.~\ref{tab:main_results_yc2_supp} and Tab.~\ref{tab:main_results_anc_supp}, our method also achieves consistent gains in $mAP$ across all the benchmarks.

Table~\ref{tab:main_results_hd_epic_supp} also reports the results of an oracle model. The oracle corresponds to training the base architectures---CG-DETR and LD-DETR, respectively--- directly on the data at the target specificity. For example, for HD-EPIC-S2, the oracle is obtained by training the baseline model on the training split of HD-EPIC-S2 derived from our proposed search-query pipeline. While this is not aligned with the underlying goal of this work, this being training on the standard captions while generalizing to more under-specified search queries, this still provides a meaningful upper-bound on the achievable performance.

Interestingly, our proposed (+SA-QD) significantly closes the gap between base and oracle model. For instance, on HD-EPIC-S2, our proposal closes the gap by up to $82\%$ of $mAP_m@0.1$.

\begin{table*}[t]
\footnotesize
\centering
\caption{Results of both CG-DETR and LD-DETR on HD-EPIC-S{
1,2,3} benchmarks with respect to our proposed modifications.} 
\label{table:comparison-results}
\resizebox{0.7\textwidth}{!}{
\begin{tabular}{c|c|c|ccc|ccc|ccc }
\toprule

\textbf{Model} & \textbf{Input} & \textbf{Variant} & \multicolumn{3}{c|}{\bm{$R_m$}} & \multicolumn{3}{c|}{\bm{$mAP_m$}} & \multicolumn{3}{c}{\bm{$mAP$}} \\
 
& & & \textbf{@0.1} &\textbf{@0.3} & \textbf{@0.5} & \textbf{@0.1} & \textbf{@0.3} & \textbf{@0.5} & \textbf{@0.1} & \textbf{@0.3} & \textbf{@0.5} \\
\midrule

\multirow{8}{*}{CG-DETR} & \multirow{2}{*}{Original} & base & 34.44 & 21.63 & 11.32 & 42.69 & 26.96 & 14.26 & 42.69 & 26.96 & 14.26 \\
& & -SA+QD & 34.24 & 22.68 & 12.59 & 45.79 & 30.52& 17.16 & 45.79 & 30.52 & 17.16 \\
\cmidrule{2-12}
& \multirow{2}{*}{S1} & base & 28.61 & 17.95 & 8.99 & 36.21 & 22.84 & 11.59 & 38.52 & 24.33 & 12.27 \\
& & -SA+QD & 29.87 & 19.69 & 10.86 & 39.74 & 26.49 & 14.87 & 41.65 & 27.98 & 15.85\\
& & base (oracle) & 28.85 & 17.44 & 9.08 & 40.42 & 25.12 & 13.1 &  42.58 & 26.74 & 14.02 \\

\cmidrule{2-12}

& \multirow{2}{*}{S2} & base & 24.71 & 15.52 & 7.89 & 32.15 & 20.1 & 10.29 &  34.19 & 21.29 & 10.79 \\
& & -SA+QD & 26.17 & 17.00 & 9.40 & 35.38 & 23.39 & 13.04 & 36.97 & 24.68 & 13.80\\
& & base (oracle) & 27.34 & 17.47 & 8.90 & 39.82 & 25.79 & 13.22 & 42.13 &  27.31 & 14.00 \\

\cmidrule{2-12}
& \multirow{2}{*}{S3} & base & 9.50 & 4.61 & 2.08 &  16.20 & 8.01 & 3.58 & 20.99  & 11.57  & 5.29 \\
& & -SA+QD & 10.57 & 6.52 & 3.45 & 17.27 & 10.65 &5.54 & 22.07  & 14.19 & 7.86 \\
& & base (oracle) & 12.31 &  6.09 & 3.06 & 23.29 &  10.94 & 5.11 & 30.56 & 15.3 & 7.37 \\

\midrule
\midrule

\multirow{8}{*}{LD-DETR} & \multirow{2}{*}{Original} & base & 34.75 & 23.90 & 13.46 & 42.59 & 29.17 & 16.42 & 42.59 & 29.17 & 16.42 \\
& & -SA+QD & 35.33 & 24.51 & 13.37 & 46.83 & 32.52 & 18.01 & 46.83 & 32.52 &  18.01 \\
\cmidrule{2-12}
& \multirow{2}{*}{S1} & base & 29.42 & 19.77 & 10.50 & 36.55 & 24.5 & 13.18 & 38.94 & 26.25 & 14.29 \\
& & -SA+QD & 30.18 & 20.26 & 10.83 & 40.50 & 27.54 & 14.94 & 42.58 & 29.15 & 16.08\\
& & base (oracle) & 29.92 & 18.61 & 8.74 & 41.5 & 26.33 & 12.78 & 43.73 & 28.00 & 13.61 \\
\cmidrule{2-12}

& \multirow{2}{*}{S2} & base & 25.23 & 16.38 & 8.46 & 32.42 & 21.11 & 10.93 & 34.32 & 22.58 & 11.89 \\
& & -SA+QD & 26.36 & 16.98 & 8.87 & 36.37 & 23.75 & 12.54 & 38.24 & 25.20 & 13.56\\
& & base (oracle) & 29.78 & 20.82 & 11.76 & 41.93 & 29.43 & 16.77 & 44.11 & 31.16 & 17.85 \\
\cmidrule{2-12}
& \multirow{2}{*}{S3} & base & 10.44 & 5.37 & 2.58 & 16.48 & 8.65 & 4.11 & 18.63 & 9.41 & 4.32\\
& & -SA+QD & 10.44 & 5.28 & 2.39 & 17.79 & 9.06 & 4.19 & 21.02 &  10.11 & 4.47\\
& & base (oracle) & 10.35 & 5.61 & 2.67 & 20.59 & 11.14 & 5.16 & 26.97 & 13.28 & 5.63 \\

\bottomrule

\end{tabular}}
\vspace{-0.2cm}
\label{tab:main_results_hd_epic_supp}
\end{table*}


\begin{table*}[h!]
\footnotesize
\centering
\caption{Results of both CG-DETR and LD-DETR on YC2-S with
respect to our proposed modification.}\label{tab:main_results_yc2_supp}
\label{table:comparison-results}
\resizebox{0.8\textwidth}{!}{
\begin{tabular}{c|c|c|cccc|cccc|cccc }
\toprule

\textbf{Model} & \textbf{Input} & \textbf{Variant} & \multicolumn{4}{c}{\bm{$R_{m}$}} & \multicolumn{4}{c}{\bm{$mAP_m$}} & \multicolumn{4}{c}{\bm{$mAP$}} \\
 
& & & \textbf{@0.1} &\textbf{@0.3} & \textbf{@0.5} & \textbf{@0.75} & \textbf{@0.1} & \textbf{@0.3} & \textbf{@0.5} & \textbf{@0.75} & \textbf{@0.1} & \textbf{@0.3} & \textbf{@0.5} & \textbf{@0.75} \\
\midrule

\multirow{4}{*}{CG-DETR} & \multirow{2}{*}{Orig.} & base & 63.32 & 50.71 & 37.79 & 20.11 & 69.47 & 55.55 & 41.04 & 17.18 & 69.47 & 55.55 & 41.04 & 17.18 \\
& & -SA+QD &  62.46 & 51.05 & 37.19 & 20.27 & 70.25 & 57.97 & 42.79 & 18.05 & 70.25 & 57.97 & 42.79 & 18.05 \\

\cmidrule{2-15}
& \multirow{2}{*}{S} & base & 28.92 & 19.87 & 11.22 & 4.60 & 38.83 & 26.96 & 15.21& 4.07 & 47.74 & 33.61 & 19.40 & 5.27 \\
& & -SA+QD & 29.97 & 20.32 & 11.38 & 4.36 & 41.00 & 29.4 & 17.21 & 4.65 &  49.52 & 36.41 & 21.86 & 6.33 \\

\midrule
\midrule

\multirow{4}{*}{LD-DETR} & \multirow{2}{*}{Orig.} & base & 68.06 & 56.34 & 39.75 & 19.69 & 73.15 & 60.62 & 42.79 & 15.88 & 73.15 & 60.62 & 42.79 & 15.88\\
& & -SA+QD & 70.20 & 55.66 & 37.06 & 17.63 & 76.35 & 61.71 & 42.01 & 15.05 & 76.35 & 61.71 & 42.01 & 15.05 \\
\cmidrule{2-15}
 & \multirow{2}{*}{S} & base & 33.13 & 23.48 & 11.70 & 4.44 & 41.69 & 30.04 & 15.58 & 4.09 & 51.86 & 37.85 & 20.0 & 5.45 \\
& & -SA+QD & 35.86 & 24.76 & 13.17 & 5.15 & 45.66 & 33.09 & 18.74 & 4.90 & 56.05 & 41.26 & 23.89 & 6.14 \\

\bottomrule

\end{tabular}}

\end{table*}

\begin{table*}[h]
\footnotesize
\centering
\caption{Results of both CG-DETR and LD-DETR on ANC-S with
respect to our proposed modification.}\label{tab:main_results_anc_supp}
\resizebox{0.8\textwidth}{!}{
\begin{tabular}{c|c|c|cccc|cccc|cccc }
\toprule

\textbf{Model} & \textbf{Input} & \textbf{Variant} &  \multicolumn{4}{c}{\bm{$R_{m}$}} & \multicolumn{4}{c}{\bm{$mAP_m$}}  & \multicolumn{4}{c}{\bm{$mAP$}} \\
 
& & & \textbf{@0.1} &\textbf{@0.3} & \textbf{@0.5} & \textbf{@0.75} & \textbf{@0.1} & \textbf{@0.3} & \textbf{@0.5} & \textbf{@0.75} & \textbf{@0.1} & \textbf{@0.3} & \textbf{@0.5} & \textbf{@0.75} \\
\midrule

\multirow{4}{*}{CG-DETR} & \multirow{2}{*}{Orig.} & base & 75.48 & 60.00 & 44.02 & 26.21 & 82.31 & 69.36 & 53.15 & 25.7 & 82.31 & 69.36 & 53.15 & 25.7 \\
& & -SA+QD & 75.19 & 60.30 & 44.96 &  26.47 &  82.25 & 69.91 & 54.5 & 25.67 & 82.25 & 69.91 & 54.5 & 25.67 \\

\cmidrule{2-15}
& \multirow{2}{*}{S} & base & 60.44 & 40.89 & 24.56 & 12.97 & 72.18 & 54.9 & 36.41 & 15.07 & 73.34 & 55.84 & 37.49 & 15.71 \\
& & -SA+QD & 63.75 & 43.12 & 25.50 & 13.36 & 74.00 & 56.42 & 37.2 & 15.09 & 75.54 & 57.52 & 38.52 & 15.78  \\

\midrule
\midrule

\multirow{4}{*}{LD-DETR} & \multirow{2}{*}{Orig.} & base & 75.72 & 60.63 & 45.17 & 27.15 & 82.73 & 70.3 & 54.46 & 26.62 & 82.73 & 70.3 & 54.46 & 26.62 \\
& & -SA+QD & 76.72 & 61.44 & 45.68 & 27.87 & 83.15 & 70.97 & 55.52 & 27.29 & 83.15 & 70.97 & 55.52 & 27.29 \\
\cmidrule{2-15}
 & \multirow{2}{*}{S} & base & 62.58 & 43.00 & 26.08 & 13.92 & 73.35 & 56.17 & 36.79 & 15.16 & 74.65 & 57.15 & 37.93 & 15.93\\
& & -SA+QD & 65.21 & 43.89 & 25.77 & 13.36 & 74.25 & 56.31 & 36.69 & 15.15 & 76.13 & 57.58 & 37.88 & 15.88 \\

\bottomrule

\end{tabular}}

\end{table*}

\subsection{Generalization across different architectures}
To verify that the performance gap is not exclusive to the DETR family, we extend our evaluation to Flash-VTG~\cite{cao2025flashvtg}, a state-of-the-art anchor-based architecture. Unlike previous models that localize via a Transformer decoder and learnable queries, Flash-VTG utilizes a \textit{Temporal Feature Layering} (TFL) module and \textit{Adaptive Score Refinement} (ASR) to regress moments from multi-scale anchors. This fundamental architectural shift allows us to test if the observed degradation is a DETR-specific artifact or a systemic challenge of the VMR task.

\noindent \textbf{Performance on under-specified queries:} We evaluated Flash-VTG on our proposed benchmarks (Table~\ref{table:comparison_flashvtg}). Despite its use of fixed anchors rather than learnable queries, Flash-VTG experiences a similar performance collapse when moving from descriptive captions to realistic search queries. On \textit{HD-EPIC-S1}, for instance, we observe a drop of $12.22$\% in mAP@0.1 (from $43.70$ to $31.48$) and $9.11$\% in R@0.1 (from $30.56$ to $21.45$). This trend is even more pronounced on \textit{YC2-S}, where mAP@0.5 plummets by $34.46$\%. These results confirm that the ``visual bias'' inherent in current datasets affects models regardless of whether they are query-based or anchor-based.

By demonstrating that the performance gap persists in a completely different architectural paradigm, we provide strong evidence that the bottleneck lies in the training data distribution rather than specific architectural choices. We leave as future work studying potential mitigation strategies for this family of methods.

\begin{table}[t]
\centering
\caption{Results of Flash-VTG on all our proposed benchmarking scenarios.}
\resizebox{1\linewidth}{!}{
\begin{tabular}{c|c|ccc|ccc}
\toprule
\textbf{Benchmark} & \textbf{Input} & \multicolumn{3}{c|}{\bm{$R_m$}} & \multicolumn{3}{c}{\bm{$mAP_m$}} \\
 
& & \textbf{@0.1} &\textbf{@0.3} & \textbf{@0.5} & \textbf{@0.1} & \textbf{@0.3} & \textbf{@0.5}\\
\midrule

\multirow{4}{*}{HD-EPIC} & \multirow{1}{*}{Original} & 30.56 & 23.83 & 16.46 & 43.70 & 36.08 & 26.89 \\
& \multirow{1}{*}{S1} & 21.45 & 16.50 & 10.98 & 31.48 & 25.71 & 18.83\\
& \multirow{1}{*}{S2} & 16.97 & 13.47 & 8.83 & 25.93 & 21.48 & 15.58 \\
& \multirow{1}{*}{S3} & 5.64 & 4.59 & 3.07 &  9.63 & 8.01 & 5.91 \\

\midrule
\multirow{2}{*}{YC2-S} & \multirow{1}{*}{Original} & 77.66 & 66.09 & 49.31 & 84.78 & 75.02 & 60.22\\
& \multirow{1}{*}{S} & 39.86 & 28.46 & 17.41 & 49.78 & 38.34 & 25.76\\

\midrule
\multirow{2}{*}{ANC-S} & \multirow{1}{*}{Original} & 74.95 & 59.12 & 43.71 & 80.93 & 68.57 & 55.26 \\
& \multirow{1}{*}{S} & 54.05 & 36.37 & 22.52 & 64.63 & 48.79 & 34.45\\

\bottomrule
\end{tabular}}
\vspace{-0.25cm}
\label{table:comparison_flashvtg}
\end{table}

%% file: sec/final_supp/ablation_realism_of_queries.tex
\section{Study of the realism of the queries}\label{sec:study_of_the_realism_of_the_queries}

\subsection{Alignment with user queries}
To ensure our generated benchmarks adequately simulate real-world search queries, we explore two different analysis:\\
\noindent \textbf{Linguistic comparison:} We perform a linguistic comparison against MS-MARCO~\cite{craswell2021ms}, a dataset of real user search logs. Our analysis shows that standard caption-based queries in HD-EPIC are significantly more descriptive than real search queries, being approximately 3x longer (17.7 vs. 5.9 words) and over-saturated with adjectives (8.1\%). Importantly, HD-EPIC-S1 queries successfully bridge this gap through our proposed under-specification pipeline. This allows, for instance, matching the MS-MARCO average length (5.98 words vs. 5.89 words) and verb density (22.6\% vs. 21\%).\\
\noindent \textbf{User realism study:} Additionally to the aforementioned quantitative analysis, we conducted a study with 22 participants who rated 20 samples each for realism. For each of the selected search queries, they were asked: \textit{``Is this search query realistic? In other words, could you imagine yourself typing this into a search bar to find this specific moment (caption provided as context)?''}. Participants gave \textit{HD-EPIC-S1} a realism score of 89\%, confirming that the queries effectively simulate actual user behavior.

\subsection{Semantic fidelity and grouping reliability}
We further validated that the under-specification process (see Sec.~\ref{sec:search_query_pipeline}) does not lose the core intent of the original captions or introduce noise into the retrieval task through three main metrics:\\
\noindent \textbf{Intent preservation:} Using dependency parsing to extract root verbs and nouns, we verified that 96.9\% of the queries in \textit{HD-EPIC-S1} preserve the original intent of the source caption. Here a search query is considered to preserve the core intent of its given caption if it contains both its main verb and noun (or a synonym). This ensures that the simplified search queries remain semantically grounded in the ground-truth video content.\\
\noindent \textbf{Discriminability:} To confirm that under-specification does not induce ambiguity, we measured the cosine similarity of the search queries against their corresponding captions using \textit{all-MiniLM-L6-v2} embeddings. The queries yielded a high similarity of $0.89$ against their own captions compared to $0.21$ against unrelated ones, providing a clear $0.68$ margin that ensures no induced confusion.\\
\noindent \textbf{Grouping reliability:} For multi-moment instances, we validate that the clusters resulting from the query grouping stage present a high intra-cluster similarity ($0.95$) and a low inter-cluster similarity ($0.31$). This indicates that only relevant captions are clustered together without ``blurring'' the boundaries with other action queries.

%% file: sec/final_supp/impact_of_calibration_in_collapse.tex
\section{Impact of calibration in the query collapse}\label{supp:impact_calibration}

\begin{figure}[t]
\centering
\includegraphics[width=0.26\textwidth]{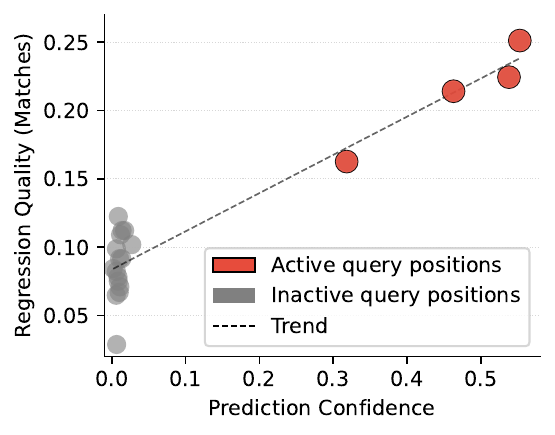}
\caption{Correlation between the average ratio of matched predictions---i.e., predictions with an IOU of at least 0.1 with one of the GT moments---with respect to their respective confidence score. This highlights the tradeoff between regression quality and confidence score quality. These results correspond to HD-EPIC-S2 for CG-DETR. \label{fig:prediction_vs_activation_correlation}.}
\end{figure}

In this section we examine if existing confidence calibration methods could resolve the query collapse issue. To this end, Fig.~\ref{fig:prediction_vs_activation_correlation} reports, for each of the learnable queries, the correlation between its regression quality---i.e., measured as the proportion of times it achieves an IOU of at least $0.1$ with a GT segment---and its confidence score. 

The plot reveals that the issue does not stem from confidence scores that fail to reflect the true quality of the regression estimate---i.e., marking as inactive, queries that in fact produce accurate predictions. Instead, what we observe is that inactive queries---i.e., with low confidence scores---produce substantially worse regression estimates. Thus, to some extent, confidence scores do capture the true quality of the regression estimate. Therefore, the core problem lies not in calibration, but in the lack of mechanisms to encourage more queries to produce accurate moment predictions.

To further support this claim, in Tab.~\ref{tab:conf_calibration} we evaluate several confidence calibration mechanisms~\cite{komisarenko2024improving, liu2022end}. These results demonstrate how these methods actually lead to a performance degradation, reinforcing that calibration alone cannot overcome active-query collapse.

\begin{table}[h]
\footnotesize
\centering
\caption{Performance of alternative confidence calibration mechanisms for CG-DETR on HD-EPIC-S2.\label{tab:conf_calibration}}
\resizebox{0.48\textwidth}{!}{
\begin{tabular}{c|ccc|ccc|c}
\toprule

\textbf{Variant} & \multicolumn{3}{c}{\bm{$R_m$}} & \multicolumn{3}{c}{\bm{$mAP_m$}} & \textbf{active}\\
 
& \textbf{@0.1} &\textbf{@0.3} & \textbf{@0.5} & \textbf{@0.1} & \textbf{@0.3} & \textbf{@0.5} & \\
\midrule
    
base & 24.71 & 15.52 & 7.89 & 32.15 & 20.1 & 10.29 & $3.64\pm1.18$ \\
\midrule
+actionness\cite{liu2022end} & 26.04 & 16.05 & 8.11 & 32.26 & 19.96 & 10.23 & $3.79\pm 0.98$ \\
+SFL\cite{komisarenko2024improving} & 21.83 & 11.22 & 4.82 & 27.54 & 14.48 & 6.31 & $3.05\pm0.99$ \\
\midrule
-SA+QD (ours) & 26.17 & 17.00 & 9.40 & 35.38 & 23.39 & 13.04 & $6.43\pm 2.16$ \\
\bottomrule
\end{tabular}}

\end{table}

%% file: sec/final_supp/extended_results_dinsentangling_single_multi.tex
\section{Extended results for disentanglement between ``single'' and ``multi''}\label{supp:extended_results_disentanglement_single_multi}

In this section we provide further details on the ablation analysis of the main text that aims evaluate separately the performance of single-moment and multi-moment queries. Concretely, in Tab.~\ref{tab:single_vs_multi_hd_epic} we disentangle the performance of single and multi-moment queries, independently, for CG-DETR evaluated on HD-EPIC-S1/S2/S3. Likewise, Tab.~\ref{tab:single_vs_multi_yc2} and Tab.~\ref{tab:single_vs_multi_anc} disentangle the performance of CG-DETR on single and multi-moment queries when evaluated on YC2-S and ANC-S benchmarks, respectively.

As discussed in the main text, even though our proposed architectural modifications generally improve in both single and multi-moment queries, we observe a more prominent improvement in the latter. This aligns with the core motivation of the proposed modifications that specifically aim to activate more decoder queries, thus being able to detect more GT moments.


\begin{table}[h]
\footnotesize
\centering
\caption{Dissection of the performance between ``single'' and ``multi'' for CG-DETR on HD-EPIC-S1/S2/S3\label{tab:single_vs_multi_hd_epic}} 
\resizebox{0.48\textwidth}{!}{
\begin{tabular}{c|c|c|ccc|ccc }
\toprule

\textbf{Input data} & \textbf{Model} & \textbf{Split} & \multicolumn{3}{c}{\bm{$R_m$}} & \multicolumn{3}{c}{\bm{$mAP_m$}} \\
 
& & & \textbf{@0.1} &\textbf{@0.3} & \textbf{@0.5} & \textbf{@0.1} & \textbf{@0.3} & \textbf{@0.5} \\
\midrule

\multirow{4}{*}{S1} & \multirow{2}{*}{base} & single & 31.47 & 20.09 & 10.19 & 39.29 & 25.02 & 12.68  \\
 & & multi & 22.90 & 13.73 & 6.77 & 29.97 & 18.46 & 9.57 \\
\cmidrule{2-9}
& \multirow{2}{*}{-SA+QD} & single & 31.61 & 21.39 & 12.25 & 41.72 & 28.23 & 16.25\\
& & multi & 26.18 & 16.16 & 8.11 & 35.66 & 22.89 & 12.12 \\
\cmidrule{1-9}

\multirow{4}{*}{S2} & \multirow{2}{*}{base} & single & 26.62 & 16.90 & 8.50 & 34.55 & 21.56 & 10.87\\
& & multi & 22.06 & 13.59 & 7.01 & 28.78 & 18.04 & 9.46\\
\cmidrule{2-9}
& \multirow{2}{*}{-SA+QD} & single & 27.02 & 18.14 & 10.34 & 36.71 & 24.66 & 13.86 \\
& & multi & 24.78 & 15.29 & 8.05 & 33.36 & 21.53 & 11.84 \\
\cmidrule{1-9}

\multirow{4}{*}{S3} & \multirow{2}{*}{base} & single & 14.66 & 8.96 & 4.06 & 21.87 & 13.03 & 6.11 \\
& & multi & 8.63 & 3.67 & 1.61 & 15.47 & 6.99 & 3.01\\
\cmidrule{2-9}
& \multirow{2}{*}{-SA+QD} & single & 15.64 & 10.36 & 6.16 & 23.23 & 15.31 & 8.88  \\
& & multi & 9.26 & 5.47 & 2.67 & 15.71 & 9.39 & 4.61 \\
  
\bottomrule

\end{tabular}}

\end{table}

\begin{table}[h]
\footnotesize
\centering
\caption{Dissection of the performance between ``single'' and ``multi'' for CG-DETR on YC2-S.\label{tab:single_vs_multi_yc2}} 
\resizebox{0.48\textwidth}{!}{
\begin{tabular}{c|c|c|ccc|ccc }
\toprule

\textbf{Input data} & \textbf{Model} & \textbf{Split} & \multicolumn{3}{c}{\bm{$R_m$}} & \multicolumn{3}{c}{\bm{$mAP_m$}}\\
 
& & & \textbf{@0.1} &\textbf{@0.3} & \textbf{@0.5} & \textbf{@0.1} & \textbf{@0.3} & \textbf{@0.5} \\
\midrule

\multirow{4}{*}{S} & \multirow{2}{*}{base} & single & 39.43 & 28.70 &  17.54 & 50.62 & 36.31 & 22.06 \\
 & & multi & 24.22 & 15.89 & 8.40 & 33.58 & 22.77  & 12.12\\
\cmidrule{2-9}
& \multirow{2}{*}{-SA+QD} & single & 39.26 & 29.47 & 17.88 & 51.83 & 39.44 & 24.75\\
& & multi & 25.82 & 16.08 & 8.48 & 36.18 & 24.81 & 13.8 \\

\bottomrule

\end{tabular}}

\end{table}

\begin{table}[h]
\footnotesize
\centering
\caption{Dissection of the performance between ``single'' and ``multi'' for CG-DETR on ANC-S.\label{tab:single_vs_multi_anc}} 
\resizebox{0.48\textwidth}{!}{
\begin{tabular}{c|c|c|ccc|ccc }
\toprule

\textbf{Input data} & \textbf{Model} & \textbf{Split} & \multicolumn{3}{c}{\bm{$R_m$}} & \multicolumn{3}{c}{\bm{$mAP_m$}} \\
 
& & & \textbf{@0.1} &\textbf{@0.3} & \textbf{@0.5} & \textbf{@0.1} & \textbf{@0.3} & \textbf{@0.5} \\
\midrule

\multirow{4}{*}{S} & \multirow{2}{*}{base} & single & 60.63 & 39.87 & 24.59 & 72.39 & 54.65 & 37.12 \\
 & & multi & 60.26 & 42.58 & 24.46 & 71.72 & 55.32 &  35.10 \\
\cmidrule{2-9}
& \multirow{2}{*}{-SA+QD} & single & 65.03  & 42.55 & 26.00 & 75.0 & 56.61 & 38.45 \\
& & multi & 61.72 & 44.16 & 24.70 & 72.42 & 56.24 & 35.21\\

\bottomrule

\end{tabular}}

\end{table}

%% file: sec/final_supp/extended_language_vs_multiplicity_gap.tex
\section{Quantifying the language vs multi-moment gap}\label{supp:quantifying_language_multi_moment_gap}

In this section, we present the full results (see Tab.~\ref{tab:extended_results_lang_multi_moment_gap}) corresponding to Fig.~\ref{fig:mult_vs_language_gap} of the main text. These results compare performance across the two given setups:$(\mathcal{D}^{captions}_{single}, \mathcal{D}^{search}_{single})$ and $(\mathcal{D}^{captions}_{multi}, \mathcal{D}^{search}_{multi})$. That is, for a given level of specificity---e.g., HD-EPIC-S2--- we first identify the subset of single-moment search queries $\mathcal{D}^{search}_{single}$. This is, the search queries that even after under-specification still map to a single GT. Then we construct the subset $\mathcal{D}^{captions}_{single}$ corresponding to the very same instances but with the caption-based query. We repeat this process for the search queries that map to multiple moments, forming the remaining subsets $\mathcal{D}^{search}_{multi}$ and $\mathcal{D}^{captions}_{multi}$.

The purpose of this construction is to evaluate the same subset---single and multi-moment instances, respectively---varying only the specificity of the textual queries. Evaluating $\mathcal{D}^{*}_{single}$ isolates the effect of the language gap, since these subsets do not contain any instance that maps to multiple moments. In contrast, $\mathcal{D}^{*}_{multi}$ evaluates the compounded effect of the language and the multi-moment gap as we are able to contrast the performance from standard caption-based queries to their corresponding search queries that contain a considerable language gap and that necessarily map to multiple moments.

\begin{table}[h]
\footnotesize
\centering
\caption{Extended results of the language vs multi-moment gap analysis from Sec.~\ref{sec:search_based_vmr}.\label{tab:extended_results_lang_multi_moment_gap}}
\label{table:comparison-results}
\resizebox{0.48\textwidth}{!}{
\begin{tabular}{c|c|c|ccc|ccc }
\toprule

\textbf{Dataset} & \textbf{Split} & \textbf{Specificity} & \multicolumn{3}{c}{\bm{$R_m$}} & \multicolumn{3}{c}{\bm{$mAP_m$}} \\
 
& & & \textbf{@0.1} &\textbf{@0.3} & \textbf{@0.5} & \textbf{@0.1} & \textbf{@0.3} & \textbf{@0.5} \\
\midrule

\multirow{4}{*}{HD-EPIC-S1} & \multirow{2}{*}{single} & Orig. & 38.53 & 24.81 & 13.04 & 46.04 & 29.74 & 15.74 \\
 & & S & 31.47 & 20.09 & 10.19 & 39.29 & 25.02 & 12.68 \\
\cmidrule{2-9}
& \multirow{2}{*}{multi} & Orig. & 25.97 & 15.03 & 7.74 & 35.75 & 21.21 & 11.2 \\
& & S & 22.90 & 13.73 & 6.77 & 29.97 & 18.46 & 9.57 \\
\midrule

\multirow{4}{*}{HD-EPIC-S2} & \multirow{2}{*}{single} & Orig. &38.79  & 25.32 & 13.26 & 46.26 & 30.25 & 16.03 \\
 & & S & 26.18 & 16.90 & 8.50 & 34.55 & 21.56 & 10.87 \\
\cmidrule{2-9}
& \multirow{2}{*}{multi} & Orig. & 28.13 & 16.26 & 8.49 & 37.5 & 22.18 & 11.68 \\
& & S & 22.06 & 13.59 & 7.01 & 28.78 & 18.04 & 9.46 \\
\midrule

\multirow{4}{*}{HD-EPIC-S3} & \multirow{2}{*}{single} & Orig. & 32.13 & 20.78 & 11.72 & 39.47 & 25.83 & 14.22 \\
 & & S & 14.66 & 8.96 & 4.06 & 21.87 & 13.03 & 6.11 \\
\cmidrule{2-9}
& \multirow{2}{*}{multi} & Orig. & 33.67 & 20.89 & 10.57 & 42.13 & 26.29 & 13.57 \\
& & S & 8.63 & 3.67 & 1.61 & 15.47 & 6.99 & 3.01 \\
\midrule

\multirow{4}{*}{YC2-S} & \multirow{2}{*}{single} & Orig. & 60.30 & 46.98 & 35.67 & 67.22 & 52.21 & 38.94 \\
 & & S & 39.43 & 28.70 & 17.54 & 50.62 & 36.31 & 22.06 \\
\cmidrule{2-9}
& \multirow{2}{*}{multi} & Orig. & 64.70 & 52.41 & 38.75 & 70.48 & 57.07 & 41.99 \\
& & S & 24.22 & 15.89 & 8.40 & 33.58 & 22.77 & 12.12 \\
\midrule

\multirow{4}{*}{ANC-S} & \multirow{2}{*}{single} & Orig. & 74.83 & 59.51 & 43.82 & 81.67 & 68.71 & 52.69 \\
 & & S & 60.63 & 39.87 & 24.59 & 72.39 & 54.65 & 37.12 \\
\cmidrule{2-9}
& \multirow{2}{*}{multi} & Orig. & 76.75 & 61.00 & 44.49 & 83.51 & 70.6 & 54.00 \\
& & S & 60.26 & 42.58 & 24.46 & 71.72 & 55.32 & 35.21 \\
\bottomrule

\end{tabular}}

\end{table}

%% file: sec/final_supp/extended_ablation_effect_of_n_decoder_queries.tex
\section{Extended ablations}\label{supp:extended_ablations}

Below we complement the ablation studies shown in Sec.~\ref{sec:experimentation} of the main text since because of space constraints, these include only the average $R_m$ and $mAP_m$ scores. Find below the corresponding results for each of the considered IOU scores. Concretely, Tab.~\ref{tab:ablation_alternative_ways_supp} shows the complete results of the main ablations regarding alternative methods to increase the number of active decoder queries. Table~\ref{tab:ablation_diversity_supp} presents the results of the ablation that studies the impact of the 1-to-1 matching strategy as a diversity promoting mechanism in our proposed (-SA+QD). Table~\ref{tab:ablation_qd_dropout_rate_supp} further details the results of the ablation that investigates the impact of the QD dropout rate. Finally, Tab.~\ref{tab:ablation_number_of_queries_supp} extends the results from the main text regarding the impact of increasing the total number of decoder queries.

Given that these findings remain consistent across benchmarks, for brevity we do not show the results for additional benchmarks.

\begin{table}[h]
\footnotesize
\centering

\caption{Extended ablation of alternative methods to increase the number of active decoder queries, evaluated with CG-DETR on HD-EPIC-S2.
\label{tab:ablation_alternative_ways_supp}}
\resizebox{0.48\textwidth}{!}{
\begin{tabular}{c|ccc|ccc }
\toprule

\textbf{Variant} & \multicolumn{3}{c}{\bm{$R_m$}} & \multicolumn{3}{c}{\bm{$mAP_m$}}\\

 & \textbf{0.1} & \textbf{0.3} & \textbf{0.5} & \textbf{0.1} & \textbf{0.3} & \textbf{0.5}\\
 
\midrule

base & 24.71 & 15.52 & 7.89 & 32.15 & 20.1 & 10.29\\
\midrule

+ 1-to-5 matching~\cite{li2022dn} & 24.00 & 15.11 & 7.97 & 29.35 & 18.36 & 9.6 \\
+ 1-to-k matching~\cite{li2022dn} & 18.31 & 8.80 & 4.12 & 18.47 & 8.87 & 4.16 \\
+group\_matching~\cite{chen2023group} & 24.33 & 15.85 & 8.86 & 31.21 & 20.12 & 11.14 \\
+hybrid matching~\cite{jia2023detrs} & 23.98 & 15.19 & 7.75 & 30.53 & 19.07 & 9.71 \\
+ms matching~\cite{zhao2024ms} & 24.13 & 15.30 & 8.02 & 31.72 & 20.35 & 10.79 \\
+data augmentation & 20.96 & 13.14 & 7.31 & 31.64 & 20.52 & 11.55 \\
\midrule
\midrule
-SA+QD (ours) & 26.17 & 17.00 & 9.40 & 35.38 & 23.39 & 13.04 \\

\bottomrule

\end{tabular}}

\end{table}

\begin{table}[h]
\footnotesize
\centering

\caption{Extended ablation of the effect of the 1-to-1 matching strategy in promoting diversity across decoder queries, for CG-DETR evaluated on HD-EPIC-S2 \label{tab:ablation_diversity_supp}} 
\resizebox{0.42\textwidth}{!}{
\begin{tabular}{c|ccc|ccc }
\toprule

\textbf{Variant} & \multicolumn{3}{c}{\bm{$R_m$}} & \multicolumn{3}{c}{\bm{$mAP_m$}}\\

 & \textbf{0.1} & \textbf{0.3} & \textbf{0.5} & \textbf{0.1} & \textbf{0.3} & \textbf{0.5}\\
 
\midrule

-SA+QD (ours) & 26.17 & 17.00 & 9.40 & 35.38 & 23.39 & 13.04 \\

\midrule

+ 1-to-k matcher~\cite{li2022dn} & 17.88 & 9.18 & 4.10 & 17.89 & 9.19 & 4.11 \\
+group matcher~\cite{chen2023group} & 25.72 & 17.17 & 9.02 & 35.36 & 23.37 & 12.41 \\
+hybrid matcher~\cite{jia2023detrs} & 26.59 & 17.60 & 9.55 & 35.79 & 24.13 & 13.24 \\

\bottomrule

\end{tabular}}

\end{table}

\begin{table}[h]
\footnotesize
\centering

\caption{Extended ablation of the impact of each of the proposed architectural modifications, for CG-DETR evaluated on HD-EPIC-S2. \label{tab:ablation_diversity_supp}} 
\resizebox{0.35\textwidth}{!}{
\begin{tabular}{cc|ccc|ccc }
\toprule
\textbf{-SA} & \textbf{+QD} & \multicolumn{3}{c}{\bm{$R_m$}} & \multicolumn{3}{c}{\bm{$mAP_m$}}\\

 & & \textbf{0.1} & \textbf{0.3} & \textbf{0.5} & \textbf{0.1} & \textbf{0.3} & \textbf{0.5}\\

\midrule
& &  24.71 & 15.52 & 7.89 & 32.15 & 20.1 & 10.29\\
\checkmark & & 23.97 & 14.73 & 7.25 & 32.17 & 20.58 & 10.33 \\
& \checkmark & 24.45 & 16.24 & 8.83 & 31.58 & 21.07 & 11.66 \\
\checkmark & \checkmark & 26.17 & 17.00 & 9.40 & 35.38 & 23.39 & 13.04 \\
\bottomrule
\end{tabular}}
\end{table}

\begin{table}[h]
\footnotesize
\centering
\caption{Extended ablation of the effect of the QD dropout rate, for CG-DETR evaluated on HD-EPIC-S2.}
\label{tab:ablation_qd_dropout_rate_supp}
\resizebox{0.35\textwidth}{!}{
\begin{tabular}{c|ccc|ccc}
\toprule
\bm{$k$} & \multicolumn{3}{c}{\bm{$R_m$}} & \multicolumn{3}{c}{\bm{$mAP_m$}}\\

 & \textbf{0.1} & \textbf{0.3} & \textbf{0.5} & \textbf{0.1} & \textbf{0.3} & \textbf{0.5}\\
\midrule
0.00 & 24.71 & 15.52 & 7.89 & 32.15 & 20.1 & 10.29\\
0.25 & 26.17 & 17.00 & 9.40 & 35.38 & 23.39 & 13.04 \\
0.50 & 1.82 & 0.86 & 0.31 & 6.72 & 3.49 & 1.32 \\
\bottomrule
\end{tabular}}
\end{table}

\begin{table}[h]
\centering
\caption{Extended ablation (corresponding to Fig.~\ref{fig:qualitative_examples} of the main text) that shows the impact of the overall number of decoder queries for the CG-DETR and our proposed (-SA + QD), for HD-EPIC-S2.}
\resizebox{0.48\textwidth}{!}{
\begin{tabular}{c|c|ccc|ccc|c}
\toprule
\textbf{Model} & \textbf{\# queries} &
\multicolumn{3}{c}{\bm{$R_m$}} &
\multicolumn{3}{c}{\bm{$mAP_m$}} &
\textbf{\# active} \\
 
 & & 0.1 & 0.3 & 0.5 & 0.1 & 0.3 & 0.5 & \\
\midrule

\multirow{4}{*}{base} & 5   & 22.41 & 14.12  & 6.94 & 27.78 & 17.46  & 8.71 & $2.87 \pm 0.70$  \\
& 10  & 23.60 & 15.39 & 7.65 & 29.4 & 19.05 & 9.59 & $3.32 \pm 1.06$  \\
& 20  & 24.71 & 15.52  & 7.89  & 32.15  & 20.1  & 10.29 & $3.52 \pm 1.05$  \\
& 30  & 21.88 & 15.00 & 7.86 & 28.83 & 19.65 & 10.39 & $3.98 \pm 1.28$  \\
& 50  & 11.66 & 6.10 & 2.74 & 18.12 & 10.02 & 4.58 & $3.66 \pm 1.10$  \\
\midrule
\multirow{4}{*}{-SA + QD (ours)} & 5  & 15.88  & 7.89 & 3.31  & 20.8  & 10.3  & 4.35 &  $2.64\pm 1.52$  \\
& 10  & 25.97 & 17.37 & 9.12 & 34.02 & 22.57 & 12.04 & $5.20\pm 1.35$ \\
& 20  & 26.17 & 17.00 & 9.40 & 35.38 & 23.39 & 13.04 & $6.73\pm 1.86$ \\
& 30  & 20.71 & 10.70 & 4.91 & 29.64 & 15.71 & 7.23 & $6.78\pm1.47$ \\
& 50  & 16.54 & 7.00  & 2.58 & 25.79 & 12.42 & 5.08 & $6.49\pm1.48$ \\
\bottomrule
\end{tabular}}

\label{tab:ablation_number_of_queries_supp}
\end{table}

%% file: sec/final_supp/qualitative_results_search_queries.tex
\section{Qualitative results of the generated search queries}\label{supp:qualitative_search_queries}
Here we show various qualitative examples of the search queries generated by our proposed search-query pipeline. More concretely, for each of the benchmarks, we first showcase various examples of the under-specifications resulting from the original caption-based queries. Then, we include various instances of final search queries, showing all the original captions that resulting from the search-query pipeline, end up mapping to the same representative search query.

\subsection{HD-EPIC}
\begin{figure}[t]
\centering
\includegraphics[width=0.48\textwidth]{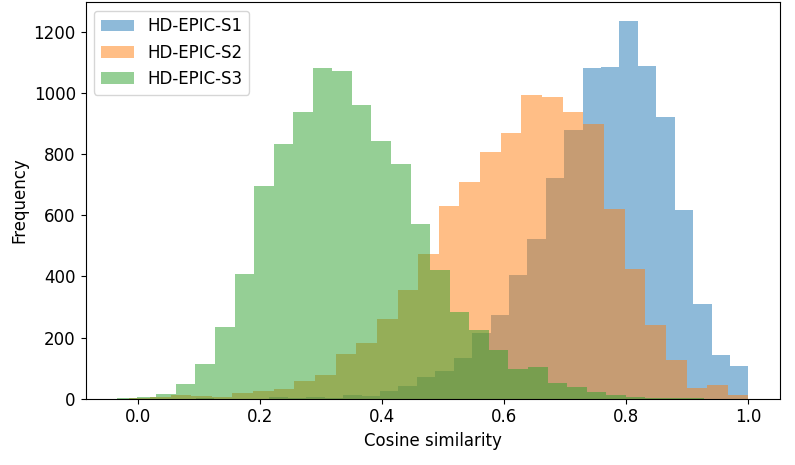}
\caption{Histogram of the feature similarities of each of the test set of each of the levels of under-specification HD-EPIC-S1/S2/S3 with respect to the original caption-based queries from HD-EPIC\label{fig:similarity_hist_hd_epic}}
\end{figure}

As explained in Sec.~\ref{supp:implementation_details}, thanks the great detail of the captions of this dataset, we were able to extract 3 different levels of under-specified search queries---S1,S2 and S3. Figure~\ref{fig:similarity_hist_hd_epic} further depicts the effect of these under-specifications by visualizing the features similarities of the under-specified search queries with respect to the original caption-based ones.

In Fig.~\ref{fig:qualitative_results_search_queries_hd_epic}, moreover, we present multiple qualitative examples of how a caption-based query progressively under-specifies in each of the proposed levels of specificity. This is further shown in Fig.~\ref{fig:final_search_queries_s1}, Fig.~\ref{fig:final_search_queries_s2} and Fig.~\ref{fig:final_search_queries_s3} where we show various final search queries, including the final search query and the multiple original captions that match it.

\begin{table}[t]
\centering
\caption{Qualitative results of the under-specified search queries for HD-EPIC-S1/S2/S3.\label{fig:qualitative_results_search_queries_hd_epic}}
\begin{tabularx}{0.9\columnwidth}{X}
\toprule
\textbf{Caption:} \textit{``Holding the ball of chicken and potato mixture in my left hand while I get some flour in my right hand so that I can sprinkle over the ball''}\\
\multicolumn{1}{c}{\textcolor{gray}{$\downarrow$}}\\[-2pt]
\textbf{S1:}\textit{``Hold the mixture and get some flour to sprinkle over it''} \\
\multicolumn{1}{c}{\textcolor{gray}{$\downarrow$}}\\[-2pt]
\textbf{S2:}\textit{``Sprinkle flour on the mixture''} \\
\multicolumn{1}{c}{\textcolor{gray}{$\downarrow$}}\\[-2pt]
\textbf{S3:}\textit{``Prepare food''} \\
\midrule
\textbf{Caption:} \textit{``Pick up tissue from inside the plate on the countertop using the right hand .''} \\
\multicolumn{1}{c}{\textcolor{gray}{$\downarrow$}}\\[-2pt]
\textbf{S1:}\textit{``Pick up tissue from the plate''} \\
\multicolumn{1}{c}{\textcolor{gray}{$\downarrow$}}\\[-2pt]
\textbf{S2:}\textit{``Pick up tissue''} \\
\multicolumn{1}{c}{\textcolor{gray}{$\downarrow$}}\\[-2pt]
\textbf{S3:}\textit{``Pick up something''} \\
\bottomrule
\end{tabularx}

\end{table}

\setlength{\tabcolsep}{2pt}
\renewcommand{\arraystretch}{1.1}
\begin{table*}[t]
\caption{Example of search queries and the captions that match it for HD-EPIC-S1. \label{fig:final_search_queries_s1}}
\begin{tabular}{p{0.95\textwidth}}
\toprule
\textbf{Representative:} \textit{``Tear the lettuce and place it on the plate''} \\
\multicolumn{1}{c}{\textcolor{gray}{$\downarrow$}}\\[-2pt]

\textbf{Caption1:} \textit{``Tear the lettuce leaves in half again and place onto the plate that is on top of the weighing scale .''} \\
\textbf{Caption2:} \textit{``Tear the lettuce leaves in half again and put the pieces onto the plate .''} \\
\textbf{Caption3:} \textit{``Use both hands to tear the lettuce leaves in half again and place the pieces onto the plate .''} \\
\textbf{Caption4:} \textit{``Use both hands to tear the lettuce leaves in half .''} \\
\textbf{Caption5:} \textit{``Use both hands to tear the lettuce leaf in half''} \\
\textbf{Caption6:} \textit{``Use the left hand to put the lettuce leaves into the bowl and then use both hands to tear the lettuce leaves in half again .''} \\
\textbf{Caption7:} \textit{``Use both hands to tear the lettuce leaves in half.''} \\
\textbf{Caption8:} \textit{``Use both hands to tear the lettuce leaves in half.''} \\
\textbf{Caption9:} \textit{``Use both hands to grip the lettuce leaves and tear them in half.''} \\
\textbf{Caption10:} \textit{``Use both hands to tear the lettuce leaves in half again.''} \\
\textbf{Caption11:} \textit{``Use both hands to tear the lettuce leaves in half and the right hand to place the leaves into the plate that is on the weighing scale.''} \\

\midrule
\textbf{Representative:} \textit{``Sprinkle flour over the chicken ball''} \\
\multicolumn{1}{c}{\textcolor{gray}{$\downarrow$}}\\[-2pt]

\textbf{Caption1:} \textit{``Using my right hand to retrieve some flour from the bowl of flour while my left hand holds the chicken ball in place so that I can sprinkle some flour over it .''} \\
\textbf{Caption2:} \textit{``Bringing my right hand closer to the potato chicken ball as I move it around in my left hand and sprinkle flour over it with my right hand .''} \\
\textbf{Caption3:} \textit{``Using my right hand to sprinkle flour over the potato chicken ball''} \\
\textbf{Caption4:} \textit{``Using my right hand to scoop some flour to sprinkle over the potato chicken ball held in my left hand so that I can use up more flour .''} \\
\textbf{Caption5:} \textit{``Placing the potato chicken ball in my left hand using my right hand so that I can sprinkle more flour over it .''} \\
\textbf{Caption6:} \textit{``Sprinkling flour over the potato chicken ball using my right hand while holding potato chicken ball in my left hand''} \\
\textbf{Caption7:} \textit{``Sprinkling flour over the potato chicken all in my left hand .''} \\

\bottomrule
\end{tabular}

\end{table*}

\setlength{\tabcolsep}{2pt}
\renewcommand{\arraystretch}{1.1}
\begin{table*}[t]
\caption{Example of search queries and the captions that match it for HD-EPIC-S2.\label{fig:final_search_queries_s2}}
\begin{tabular}{p{0.95\textwidth}}
\toprule
\textbf{Representative:} \textit{``Close a cupboard''} \\
\multicolumn{1}{l}{\textcolor{gray}{$\quad \quad \downarrow$}}\\[-2pt]
\textbf{Caption1:} \textit{``With my right hand , close the bottom cupboard .''} \\
\textbf{Caption2:} \textit{``With my right hand , close the cupboard .''} \\
\textbf{Caption3:} \textit{``Close the cupboard by pushing it with my left hand .''} \\
\textbf{Caption4:} \textit{``Having realized this is the wrong cupboard , close it again by pushing it with my left hand .''} \\
\textbf{Caption5:} \textit{``I close the cupboard by pushing it with my right hand .''} \\
\textbf{Caption6:} \textit{``Close the cupboard with my right hand .''} \\
\textbf{Caption7:} \textit{``Close the cupboard by pushing it with my left hand .''} \\
\textbf{Caption8:} \textit{``With my right hand , close the cupboard by pushing it .''} \\
\textbf{Caption9:} \textit{``Not finding what I was looking for , close again the cupboard with my left hand .''} \\
\textbf{Caption10:} \textit{``With my left hand , close the cupboard by pulling it towards me .''} \\
\textbf{Caption11:} \textit{``With my right hand , close the smaller cupboard on the right hand side .''} \\
\textbf{Caption12:} \textit{``With my right hand , close the cupboard by pulling the cupboard towards me .''} \\
\textbf{Caption13:} \textit{``Close the cupboard using my left hand .''} \\
\textbf{Caption14:} \textit{``With my right hand , close the cupboard by pulling the cupboard .''} \\
\textbf{Caption15:} \textit{``Using my left hand , close the other larger cupboard .''} \\
\textbf{Caption16:} \textit{``Close the small cupboard using my right hand .''} \\
\textbf{Caption17:} \textit{``With my right hand , close the top cupboard .''} \\
\textbf{Caption18:} \textit{``With my right hand , close the bottom cupboard .''} \\
\textbf{Caption19:} \textit{``With my left hand , close the cupboard in the corner next to the dishwasher by pushing the cupboard .''} \\

\midrule
\textbf{Representative:} \textit{``Open a lid''} \\
\multicolumn{1}{c}{\textcolor{gray}{$\downarrow$}}\\[-2pt]
\textbf{Caption1:} \textit{``Open the lid of the trash bin by flipping it .''} \\
\textbf{Caption2:} \textit{``Open the trash bin 's lid .''} \\
\textbf{Caption3:} \textit{``Open the lid of the trash bin . This action occurs off the screen .''} \\
\textbf{Caption4:} \textit{``Open the lid of the food waste bin .''} \\

\bottomrule
\end{tabular}

\end{table*}

\setlength{\tabcolsep}{2pt}
\renewcommand{\arraystretch}{1.1}
\begin{table*}[t]
\caption{Example of search queries and the captions that match it for HD-EPIC-S3. \label{fig:final_search_queries_s3}}
\begin{tabular}{p{0.95\textwidth}}
\toprule
\textbf{Representative:} \textit{``Open a book''} \\
\multicolumn{1}{c}{\textcolor{gray}{$\downarrow$}} \\[-2pt]

\textbf{Caption1:} \textit{``Push down in the center of the recipe book along the spine to try and make sure it stays open onto the countertop.''} \\
\textbf{Caption2:} \textit{``Pick up the recipe book using the right hand and flipping it over so that the recipe can be seen.''} \\
\midrule
\textbf{Representative:} \textit{``Clean a bowl.''} \\
\multicolumn{1}{c}{\textcolor{gray}{$\downarrow$}} \\[-2pt]
\textbf{Caption1:}   \textit{``Realise there is some old food or dirt on the bottom of the bowl.''} \\
\textbf{Caption2:}   \textit{``Try to scratch away the food or dirt to see if it's on the bottom of the bowl or inside the bowl using the right hand.''} \\
\textbf{Caption3:}   \textit{``Clean the bottom of the bowl using the kitchen roll in the right hand whilst holding the bowl down on the left hand.''} \\
\textbf{Caption4:}   \textit{``Grab hold of the sponge using the right hand and continue to clean the inside of the large mixing bowl.''} \\
\textbf{Caption5:}   \textit{``Submerge the bowl into the water and clean the inside of the bowl using the sponge in the right hand, paying close attention to get rid of any food that has been stuck onto the bowl.''} \\
\textbf{Caption6:}   \textit{``Rotate the bowl over to clean the inside once more.''} \\
\bottomrule
\end{tabular}

\end{table*}

\subsection{YC2-S}
\begin{figure}[t]
\centering
\includegraphics[width=0.48\textwidth]{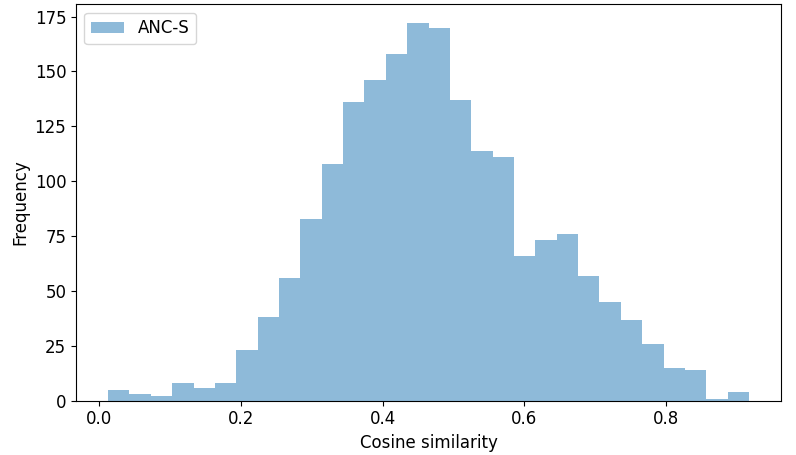}
\caption{Histogram of the feature similarities of the test set of YC2-S with respect to the original caption-based queries from YC2\label{fig:similarity_hist_yc2}}
\end{figure}

Below we repeat the same analysis for YC2-S benchmark. Concretely, find in Fig.~\ref{fig:similarity_hist_yc2} the histogram of the features similarities between the original caption-based queries and the corresponding search queries. Additionally, Fig.~\ref{fig:qualitative_results_search_queries_yc2} provides various qualitative examples of under-specifications of original captions from YC2~\cite{zhou2018towards} into their corresponding search queries, while Fig.~\ref{fig:final_search_queries_yc2_s} shows various final search queries.

\begin{table*}[t]
\centering
\caption{Qualitative results of the under-specified search queries for YC2-S.\label{fig:qualitative_results_search_queries_yc2}}
\setlength{\tabcolsep}{2pt}
\renewcommand{\arraystretch}{1}
\begin{tabularx}{0.9\columnwidth}{X}
\toprule
\textbf{Caption:} \textit{``Add strained tomato puree to the blond roux and stir the mixture continuously''} \\
\multicolumn{1}{c}{\textcolor{gray}{$\downarrow$}}\\[-2pt]
\textbf{S:}\textit{``Add ingredient.''} \\
\midrule
\textbf{Caption:} \textit{``Add salt to the pan and mix''} \\
\multicolumn{1}{c}{\textcolor{gray}{$\downarrow$}}\\[-2pt]
\textbf{S:}\textit{``Season food.''} \\

\bottomrule
\end{tabularx}

\end{table*}

\setlength{\tabcolsep}{2pt}
\renewcommand{\arraystretch}{1.1}
\begin{table*}[t]
\caption{Example of search queries and the captions that match it for YC2-S.\label{fig:final_search_queries_yc2_s}}
\begin{tabular}{p{0.95\textwidth}}
\toprule
\textbf{Representative:} \textit{``Add ingredients.''} \\
\multicolumn{1}{c}{\textcolor{gray}{$\downarrow$}}\\[-2pt]

\textbf{Caption1:} \textit{``Crack one egg into a bowl''} \\
\textbf{Caption2:} \textit{``Add one table spoon of oil salt and cayenne pepper and baking powder and beat''} \\
\textbf{Caption3:} \textit{``Add one cup of beer and mix''} \\
\textbf{Caption4:} \textit{``Add one quarter cup of corn meal and one cup of flour''} \\
\textbf{Caption5:} \textit{``Add onions into batter and drop into hot oil''} \\

\midrule
\textbf{Representative:} \textit{``Cook potatoes.''} \\
\multicolumn{1}{c}{\textcolor{gray}{$\downarrow$}} \\[-2pt]
\textbf{Caption1:} \textit{``Spread rock salt on a baking tray place potatoes on it and draw few spikes.''} \\
\textbf{Caption2:} \textit{``Pierce the knife inside the potatoes and find if the potatoes are cooked properly.''} \\
\textbf{Caption3:} \textit{``Cut the cooked potatoes in half and scoop the flesh and put it in a bowl.''} \\
\textbf{Caption4:} \textit{``Keep the mashed potatoes on the flame and mix adding butter until it is mix well and the bottom of the pan becomes shiny.''} \\

\bottomrule
\end{tabular}

\end{table*}

\subsection{ANC-S}
\begin{figure}[t]
\centering
\includegraphics[width=0.48\textwidth]{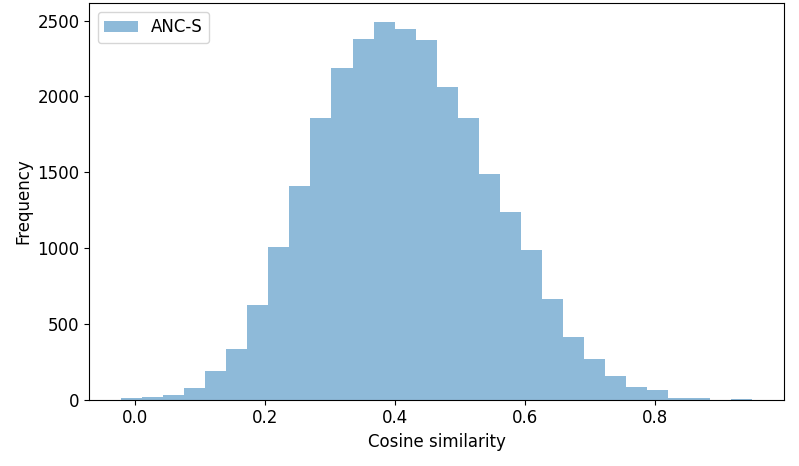}
\caption{Histogram of the feature similarities of each of the test set of ANC-S with respect to the original caption-based queries from ANC\label{fig:similarity_hist_anc}}
\end{figure}

Finally, we perform the same analysis for ANC-S benchmark. Concretely, find in Fig.~\ref{fig:similarity_hist_anc} the histogram of the features similarities between the original caption-based queries and the corresponding search queries. Additionally, Fig.~\ref{fig:qualitative_results_search_queries_anc_s} provides various qualitative examples of under-specifications of original captions from ANC~\cite{krishna2017dense} into their corresponding search queries, while Fig.~\ref{fig:final_search_queries_anc_s} shows various final search queries.

\begin{table*}
\centering
\caption{Qualitative results of the under-specified search queries for ANC-S.\label{fig:qualitative_results_search_queries_anc_s}}
\setlength{\tabcolsep}{2pt}
\renewcommand{\arraystretch}{1.1}
\begin{tabularx}{\columnwidth}{X}
\toprule
\textbf{Caption:} \textit{``The man is washing a side of the car.''} \\
\multicolumn{1}{c}{\textcolor{gray}{$\downarrow$}} \\[-2pt]
\textbf{S:} \textit{`A person performs a task on a vehicle.''} \\
\midrule
\textbf{Caption:} \textit{``The person then moves back and fourth on the machine while rowing his arms back and fourth.''} \\
\multicolumn{1}{c}{\textcolor{gray}{$\downarrow$}} \\[-2pt]
\textbf{S:} \textit{``A person uses a machine.''} \\
\bottomrule
\end{tabularx}

\end{table*}

\setlength{\tabcolsep}{2pt}
\renewcommand{\arraystretch}{1.1}
\begin{table*}[t]
\caption{Example of search queries and the captions that match it for ANC-S.\label{fig:final_search_queries_anc_s}}
\begin{tabular}{p{0.95\textwidth}}
\toprule
\textbf{Representative:} \textit{``People play a game.''} \\
\multicolumn{1}{c}{\textcolor{gray}{$\downarrow$}} \\[-2pt]

\textbf{Caption1:} \textit{``A man and woman are shown standing on a tennis court passing a ball back and fourth.''} \\
\textbf{Caption2:} \textit{``The players volley and the play is pressed to the back of the court hitting long shots.''} \\
\textbf{Caption3:} \textit{``The players volley and the birdie is hit out of view and the player retrieves it then serves.''} \\
\textbf{Caption4:} \textit{``The players have a long volley until the play in the foreground misses and the birdie lands at her feet.''} \\
\textbf{Caption5:} \textit{``People play games of badminton on indoor courts.''} \\
\textbf{Caption6:} \textit{``Behind them are another group of people playing and the man and woman continuously pass the ball back ad fourth to one another.''} \\
\midrule
\textbf{Representative:} People perform household chores. \\
\multicolumn{1}{c}{\textcolor{gray}{$\downarrow$}} \\[-2pt]

\textbf{Caption1:} \textit{``A young girl and boy are washing dishes in a kitchen.''} \\
\textbf{Caption2:} \textit{``They are doing the dishes in the sink.''} \\
\textbf{Caption3:} \textit{``the mother enters and adds some dishes from the rack back into the sink to be rinsed again and shows the boy what was wrong with the pot.''} \\
\bottomrule
\end{tabular}

\end{table*}